\documentclass[10pt]{article} 

\usepackage[accepted]{rlj}
\usepackage{definition}
\usepackage{titletoc}
\usepackage{booktabs}
\tcbuselibrary{breakable=false,skins}

\usepackage{amssymb}            
\usepackage{mathtools}          
\usepackage{mathrsfs}           
\usepackage{graphicx}           
\usepackage{subcaption}         
\usepackage[space]{grffile}     
\usepackage{url}                
\usepackage{lipsum}             

\title{Reward Design Agent for Reinforcement Learning}

\setrunningtitle{RDA: Reward Design Agent for Reinforcement Learning}

\author{Hojoon Lee\textsuperscript{1,2,$\dagger$}, 
Ajay Subramanian\textsuperscript{1,3,$\dagger$}, 
Ben Abbatematteo\textsuperscript{1}, \\
Vijay Veerabadran\textsuperscript{1},
Pedro Matias\textsuperscript{1},
Karl Ridgeway\textsuperscript{1},
Nitin Kamra\textsuperscript{1}
}

\emails{hojoon.lee@holiday-robotics.com, nitinkamra@meta.com}

\affiliations{
$^{1}$\textbf{Meta}
$^{2}$\textbf{Holiday Robotics}
$^{3}$\textbf{Boston Dynamics}
\par 
$^\dagger$ Work done during internship at Meta
}

\contribution{
    We empirically show that LLM-based automated reward generation can produce policies that achieve high task success while misaligned with natural-language task instructions, as revealed by vision–language–model–based trajectory evaluation.
    }
    {
    Prior work on LLM-based reward generation demonstrates strong performance on relatively simple locomotion and manipulation tasks~\cite{ma2023eureka,ma2024dreureka}. As task complexity increases, however, task completion alone does not ensure adherence to the full semantics of the instructions.
    }
\contribution{
    We introduce the Reward Design Agent (RDA), a vision–language–model–based agentic framework that combines several components for improving instruction alignment, including instruction decomposition into subtasks, visual evaluation of policy trajectories, subtask-level misalignment identification, and iterative reward code revision.
    }
    {
    Prior LLM-based reward generation methods rely largely on numerical feedback~\cite{ma2023eureka} and lack explicit mechanisms for visual trajectory understanding or structured reasoning over instruction semantics.
    }
\contribution{
    Across 12 ManiSkill tabletop manipulation tasks~\cite{tao2024maniskill3} and 4 HumanoidBench whole-body locomotion tasks~\cite{sferrazza2024humanoidbench}, RDA produces highly instruction-aligned policies under GPT-4.1 based VLM evaluation, while achieving task success rates competitive with human-designed rewards and prior automated reward generation methods.
    }
    {
    These benchmarks include challenging tasks such as charger insertion and whole-body package delivery, which are difficult for both manually designed rewards and prior automated reward design approaches. On some tasks, RDA's pure task success rate is lower than Eureka's; our central contribution concerns alignment quality rather than raw success.
    }


\summary{
Hand-designed reward functions have driven recent advances in reinforcement learning for robotics and games, but they are expensive to construct, difficult to tune, and frequently misaligned with the designer's intent. Recent large language model-based reward generation methods, such as Eureka~\cite{ma2023eureka}, reduce manual effort by generating and refining reward code from natural language task instructions. However, these methods typically rely on coarse optimization signals such as success rates or reward statistics, without explicitly analyzing policy behavior. As a result, reward learning can suffer from instruction misalignment, where policies may complete the nominal task objective by exploiting unintended behaviors that diverge from the instruction's full semantics.

We introduce the Reward Design Agent (RDA), an agentic framework for reward design that explicitly targets alignment by incorporating semantic feedback at the trajectory level using vision language models. RDA combines several components for grounding reward refinement in behavioral semantics rather than scalar metrics alone: decomposing natural language instructions into structured subtasks, evaluating policy trajectories through visual understanding to assess subtask completion, identifying behavioral failure modes, and revising reward code accordingly. On 12 ManiSkill tabletop manipulation tasks and 4 HumanoidBench whole-body locomotion tasks, RDA produces highly instruction-aligned policies under GPT-4.1-based VLM evaluation while achieving task success rates competitive with human-designed rewards and prior automated reward design methods.
}

\begin{document}

\maketitle  

\begin{abstract}

Reinforcement learning has enabled the acquisition of impressive robotic skills, but typically requires hand-crafted reward functions that are slow to design and difficult to align with human intentions.
Recent work, such as Eureka, automates reward design by using an LLM to iteratively generate and refine reward code from task descriptions. However, they rely on coarse feedback signals such as success rate, which provide little semantic insight into the learned behavior. As a result, their trained policies achieve the final goal but are frequently poorly aligned with task instructions.
We introduce the Reward Design Agent (RDA), a VLM-based agentic framework that injects semantic understanding into reward design. 
RDA decomposes tasks, visually evaluates trajectories, summarizes failure modes, and iteratively revises reward code to better align with task instructions. Across 12 tabletop manipulation tasks from ManiSkill and 4 whole-body manipulation tasks from HumanoidBench, RDA produces policies substantially more instruction-aligned than those of other baselines, while achieving comparable task success rates. Videos and generated reward code are available on our \href{https://nitinkamra1992.github.io/reward-design-agent/}{project page}.

\end{abstract}


\section{Introduction}
\label{section:intro}

Reinforcement learning (RL) combined with highly-parallelizable simulators~\citep{todorov2012mujoco,makoviychuk2021isaac,tao2024maniskill3} has demonstrated impressive successes in robotic control, from quadrupeds traversing rough terrain to humanoids performing dexterous manipulation~\citep{hwangbo2019learning, andrychowicz2020learning, tang8deep}. 
These successes rely on hand-crafted reward functions that must be manually designed and tuned per task—a process that is slow, brittle, and dominated by trial-and-error~\citep{booth2023perils}, limiting the scalability of RL.

\begin{figure}[t]
\begin{center}
\includegraphics[width=.9\textwidth]{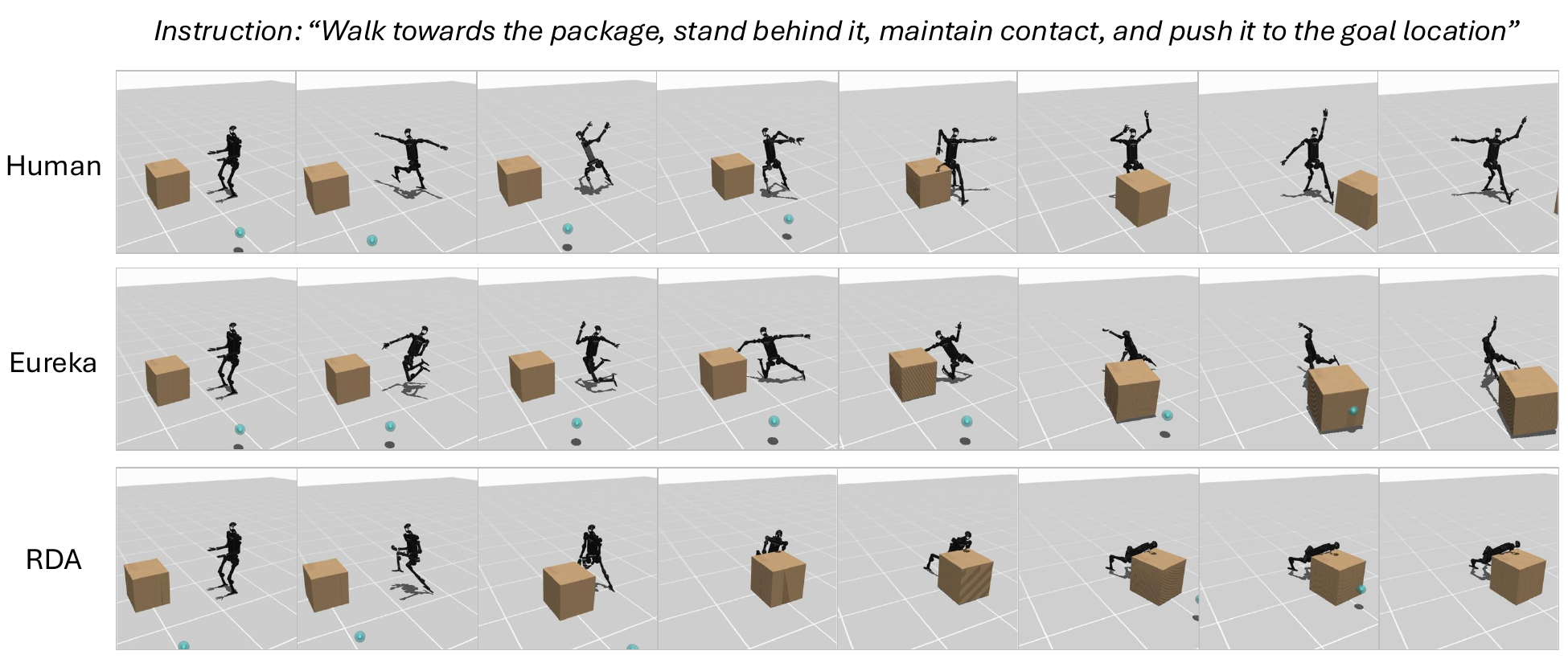}
\end{center}
\vspace{-4mm}
\caption{\textbf{Human vs Eureka vs RDA.}
The intended behavior is to approach the package, stand behind it, maintain contact, and gently push it to the target.
Rewards designed by both humans and Eureka achieve task success but induce misaligned behaviors, causing the humanoid to kick or throw the package rather than push it smoothly.
In contrast, RDA produces a policy that succeeds in the task while following the intended behavior.}

\vspace{-2mm}
\label{figure:humanoid_package}
\end{figure}

Reward design is intrinsically difficult because a single scalar must encode multiple, often competing objectives~\citep{singh2009rewards}. Consider a humanoid instructed to push a heavy object to a target: it must walk toward the object while balancing, establish contact, and push smoothly. Over-weighting any objective destabilizes behavior, and subtle mis-specifications can induce misaligned policies~\citep{knox2023reward, booth2023perils}: a reward that mainly rewards the object reaching the goal may lead the agent to throw it rather than deliver it safely. As shown in Figure~\ref{figure:humanoid_package}, a human-designed reward gets the humanoid to the box but fails to complete delivery. As tasks grow more complex and temporally extended, manual reward design becomes impractical.

To reduce this burden, a large body of work learns rewards from data. Inverse RL infers rewards from expert demonstrations~\citep{ng2000algorithms, abbeel2004apprenticeship, ziebart2008maximum}, while preference-based methods learn from human feedback~\citep{christiano2017deep, lee2021pebble, hejna2023few} or from comparisons produced by vision-language models (VLMs)~\citep{yang2024trajectory, wang2024rl_vlmf, luu2025erlvlm}. Other approaches estimate task progress from demonstrations~\citep{ma2022vip, hung2025victor, zhang2025rewind} or leverage VLM priors via in-context learning~\citep{ma2024gvl} or fine-tuning~\citep{zhai2025vlac, lee2026roboreward}. However, these methods typically depend on large, expensive reward models that are costly to evaluate in highly parallelized simulators and hard to adapt when policies misbehave.

A complementary line of work instead encodes human intention directly into natural-language instructions, prompting foundation models to generate executable reward code over simulator state. Such code is lightweight, runs efficiently in massively parallel simulators, and remains interpretable. Eureka~\citep{ma2023eureka} introduced this paradigm, casting reward design as evolutionary search over LLM-generated candidates: the LLM proposes reward functions, agents are trained on each, the best are selected, and the LLM refines them using per-component reward statistics. A follow-up~\citep{ma2024dreureka} shows that safety-aware prompting reduces reward hackability. On standard locomotion and manipulation tasks, these methods match or surpass expert-designed rewards.

Despite this promise, Eureka frequently produces misaligned rewards because its reflection loop lacks visual assessment and relies solely on coarse reward statistics. Recent work shows that evaluating intent–behavior alignment requires visual, trajectory-level analysis~\citep{muslimani2025towards}. Because Eureka ignores induced trajectories, distinct failure modes appear numerically identical, obscuring how to revise the reward code. In Figure~\ref{figure:humanoid_package}, for example, Eureka solves package delivery by throwing the package rather than pushing it, and fails to diagnose or repair this misalignment.

We introduce Reward Design Agent (RDA), a VLM-based agentic framework that mirrors how humans design rewards by (1) decomposing complex tasks into subtasks, (2) interpreting failures at the subtask level, and (3) reflecting in a targeted manner. RDA parses the natural-language instruction into subtasks and proposes reward candidates conditioned on them. It then runs a closed-loop evolutionary process: it trains RL policies on the candidate rewards, visually analyzes the resulting trajectories to score subtask completion, diagnoses failure modes, and retains the best candidates. Guided by these diagnostics, RDA revises both the subtask specifications and the reward code before the next iteration, enabling targeted corrections and progressively improving reward quality.

We evaluate RDA on 12 short-horizon tabletop manipulation tasks from ManiSkill~\citep{tao2024maniskill3} and 4 long-horizon humanoid whole-body manipulation tasks from HumanoidBench~\citep{sferrazza2024humanoidbench}. 
On ManiSkill, both Eureka and RDA outperform human-designed rewards in terms of success rate while maintaining high task alignment; however, only RDA solves the most challenging task (PlugCharger). On HumanoidBench, both methods again surpass human rewards but diverge qualitatively. As shown in Figure~\ref{figure:humanoid_package}, Eureka violates task specifications (e.g., throwing packages instead of delivering them) while RDA follows instructions correctly.
Taken together, these results demonstrate that RDA generates performant and instruction-aligned rewards through an agentic loop that decomposes tasks, diagnoses failures, and revises rewards accordingly.

\section{Related Work}
\label{sec:related}

\subsection{Learning Reward Functions from Data}

A large body of work replaces hand-crafted rewards by inferring them from data. Classical inverse RL assumes expert demonstrations are approximately optimal and infers a reward under which they are high-likelihood or high-return~\citep{ng2000algorithms,abbeel2004apprenticeship,ziebart2008maximum}. Adversarial formulations instead cast imitation as distribution matching~\citep{ho2016gail,peng2018variational,kostrikov2018discriminator}, with later work using it as an auxiliary shaping term rather than the full objective~\citep{peng2021amp,tessler2024maskedmimic}.

Another line of work learns rewards from preferences. Rather than relying on expert trajectories, users compare pairs of trajectories~\citep{christiano2017deep,lee2021pebble,hejna2023few} or provide richer forms of feedback~\citep{wilde2022learning,myers2022learning}. To reduce human annotation effort, recent approaches replace human evaluators with vision–language models (VLMs) that automatically generate preference labels~\citep{yang2024trajectory,wang2024rl_vlmf,luu2025erlvlm,luu2025policy}, though such supervision is often sparse and insufficient for fine-grained control. Orthogonally, progress-based reward proxies can be learned from demonstrations~\citep{ma2022vip,hung2025victor,zhang2025rewind}, with recent work further leveraging VLM priors through in-context learning~\citep{ma2024gvl} or fine-tuning~\citep{zhai2025vlac, lee2026roboreward}.

Despite reducing manual engineering, these approaches still require curated datasets, large reward models that are costly to evaluate in parallel simulators, and are hard to correct once misaligned.

\subsection{Generating Rewards from Foundation Models}

Rather than learning reward models from data, recent alternatives directly query foundation models for scalar feedback. Vision–language encoders such as CLIP score trajectories via video–text similarity~\citep{sontakke2023roboclip,rocamonde2023vlmrl,baumli2023vlmsource}, and frozen video models provide rewards as conditional likelihoods~\citep{escontrela2023viper}. LLM-based approaches instead produce textual rationales and score them against goal instructions~\citep{kwon2023rdllm,hu2023language}.
These signals offer global feedback without reward-model training, but remain coarse: similarity collapses distinct failure modes and overlooks the fine-grained, step-by-step information needed to guide low-level control.


\subsection{Generating Reward Code}

In simulators, reward functions are conventionally written as executable code over privileged state~\citep{hwangbo2019learning,andrychowicz2020learning}. To reduce design burden, early evolutionary approaches optimized parameters of hand-designed templates~\citep{niekum2010genetic,faust2019evolving,chiang2019learning}. More recently, LLM-based methods synthesize reward code directly over simulator state~\citep{yu2023language}. Eureka~\citep{ma2023eureka} embeds an LLM in an evolutionary loop: the LLM proposes a candidate reward code, policies are trained, top candidates are selected, and the LLM reflects on reward statistics to refine the next generation. A subsequent extension combines real-world experiments~\citep{nahrendra2026locovlm} with safety-aware prompting~\citep{ma2024dreureka}.

However, Eureka's reflection signal is purely numerical (returns and component statistics) and coarse. Distinct failure modes with similar scores are indistinguishable, preventing the system from diagnosing why a reward failed or how to fix it. Recent work has incorporated visual signals on top of Eureka: using VLMs to filter unsafe candidates~\citep{ma2025automated} or adding CLIP-based alignment rewards~\citep{cui2025grove}. While these approaches detect misalignment, they lack actionable diagnostics for reward revision. Concurrently, Prompt-to-Policy~\citep{prompt2policy2026} automates the reward-engineering loop with role-specialized agents and VLM-based trajectory judging.

In contrast, RDA uses a VLM to generate subtask-level diagnostics that directly provide targeted edits, closing the loop between detecting and correcting misalignment.

\begin{figure}[t!]
\begin{center}
\includegraphics[width=.99\textwidth]{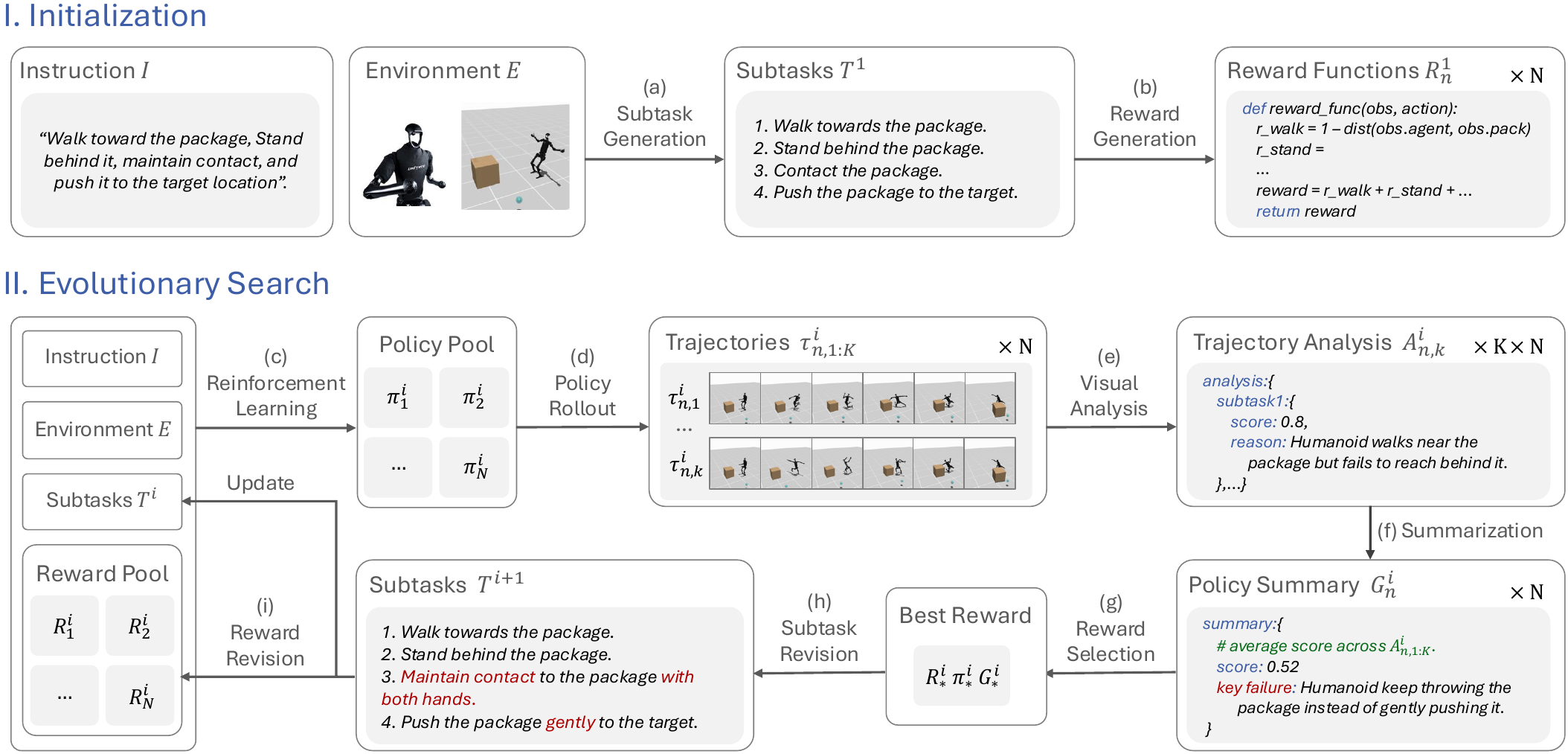}
\end{center}
\vspace{-2mm}
\caption{\textbf{RDA Overview.} 
\textbf{(I) Initialization.} Given a natural-language instruction $I$ and a simulator environment $E$, RDA first decomposes $I$ into a list of subtasks $T^1$. Conditioned on $(I,E,T^1)$, it then generates an initial pool of reward candidates $\{R^1_1,\dots,R^1_N\}$ to seed the search process.
\textbf{(II) Evolutionary Search.} During the evolutionary loop, for each iteration $i$, each reward function $R^i_n$ is used to train a policy $\pi^i_n$ in simulation. 
$K$ trajectories are sampled from each policy: $\tau^i_{n,1:K}$ and visually evaluated by a VLM, which produces an analysis report $A^i_{n,k}$, containing subtask-wise scores and evaluation rationales.
These reports are aggregated across trajectories into a summary $G_n^i$, which is then used to revise both the subtask list and the reward code. This process repeats for a fixed number of iterations.
}
\vspace{-1mm}
\label{figure:overview}
\end{figure}

\section{Preliminary}

We formalize reward design as the problem of selecting a reward function that induces a learnt policy whose behavior satisfies a target instruction.
A designer specifies the intended behavior via an instruction $I \in \mathcal{I}$ (e.g., text, video, or other modalities). The system’s goal is to produce a reward function $R \in \mathcal{R}$ such that the policy $\pi \in \Pi$ returned by a learning procedure is aligned with $I$.
Building on the formulation of \citep{singh2009rewards}, we extend to instruction-conditioned settings and provide a formal definition as follows:

\begin{definition}[Reward Design Problem (RDP)]
The RDP consists of a tuple
\[
P \;=\; \bigl\langle \mathcal{I}, \mathcal{E}, \mathcal{R}, \mathcal{L}, \mathcal{S} \bigr\rangle
\]
where $\mathcal{I}$ is the instruction space, $\mathcal{E}$ is the environment space and $\mathcal{E}=\langle \mathcal{O}, \mathcal{A}, \mathcal{T} \rangle$ represents observation space $\mathcal{O}$, action space $\mathcal{A}$, and transition dynamics $\mathcal{T}$ respectively. 
$\mathcal{R}$ is the reward function space.
$\mathcal{L} : \mathcal{E} \times \mathcal{R} \to \Pi$ is a learning algorithm mapping environment and reward function to policy space $\Pi$.
$\mathcal{S} : \Pi \times \mathcal{I} \to \mathbb{R}$ is a score function that evaluates instruction alignment by comparing a policy's induced behavior to an instruction.

Given an instruction $I \in \mathcal{I}$, an environment $E \in \mathcal{E}$, learning algorithm $L \in \mathcal{L}$, and score function $S \in \mathcal{S}$, the objective of RDP is to find a reward function $R^{\star} \in \mathcal{R}$ that maximizes the score funcion:
\[
R^\star \;\in\; \arg\max_{R \in \mathcal{R}} \;  S\bigl(L(E, R),\, I\bigr).
\]
\end{definition}

\section{RDA: Reward Design Agent}

We propose RDA as an automated algorithm for solving RDP.
Given a RDP, $P=\langle \mathcal{I},\mathcal{E},\mathcal{R},\mathcal{L},\mathcal{S}\rangle$, we assume the learning procedure $\mathcal{L}$ and the score function $\mathcal{S}$ are fixed. RDA performs an evolutionary search over $\mathcal{R}$ and employs a VLM to visually analyze policies and refine the reward function.

\subsection{Initialization}
\vspace{-1mm}

\textbf{Input Preparation.} As illustrated in the top-left side of Figure~\ref{figure:overview}, RDA takes as input a natural-language instruction $I \in \mathcal{I}$ that specifies the intended behavior (e.g., “with natural motion, walk to the package and push it with both hands”), and a simulated environment $E \in \mathcal{E}$.

\textbf{Subtask Generation.}
Given $(I,E)$, the VLM decomposes the instruction into a sequence of $J$ interpretable subtasks $T^1 = [t_1^1,\ldots,t_{J}^1]$ (Figure~\ref{figure:overview}(a)). This structured representation enables fine-grained reasoning about which specific aspects of the desired behavior are satisfied or violated.

\textbf{Reward Generation.} 
Conditioned on $(I, E, T^1)$, the VLM generates an $N$ reward candidates:
$$\{R_1^1, \ldots, R_N^1\} \subset \mathcal{R}.$$
Each candidate $R_n^1$ is executable code that maps simulator states to scalar rewards. Rewards are structured as weighted sums of subtask-level components, where component $r_j$ corresponds to subtask $t_j^1$. All candidates are compiled and validated before proceeding to the evolutionary loop.

\subsection{Evolutionary Search}
\vspace{-1mm}

RDA performs an evolutionary search for $M$ iterations. At iteration $i \in \{1, \ldots, M\}$, we maintain $N$ reward candidates $\{R_n^i\}_{n=1}^N$ and their associated subtask decomposition $T^i = [t_j^i]_{j=1}^{J}$. We use superscript $i$ for the iteration index and subscript $n$ for the candidate index.

\textbf{Policy training.} For each reward candidate $R_n^i$, we train a policy via:
$$\pi_n^i = \mathcal{L}(E, R_n^i),$$
using reinforcement learning in the simulator (Figure~\ref{figure:overview}(c)). To reduce computational cost, training may be warm-started from the best checkpoint of the previous iteration.

\textbf{Policy Rollout.} Each policy $\pi_n^i$ is evaluated over $K$ episodes in $E$, yielding a set of trajectories:
$$\{\tau_{n,k}^i\}_{k=1}^K,$$
where $\tau_{n,k}^i$ denotes the $k$-th trajectory for candidate $n$ at iteration $i$. For each trajectory, we record the rendered video and subtask-level reward logs (Figure~\ref{figure:overview}(d)).

\textbf{Visual Analysis.}
For each trajectory–subtask pair $(\tau_{n,k}^i, t_j^i)$, the VLM evaluates whether the observed behavior satisfies subtask $t_j^i$. The VLM returns a completion score $s_{n,k,j}^i \in [0,1]$ indicating the degree to which subtask $t_j^i$ is satisfied, and a language description $\rho_{n,k,j}^i$ explaining the score. We denote the complete analysis for trajectory $\tau_{n,k}^i$ as:
$$A_{n,k}^i = \{(s_{n,k,j}^i, \rho_{n,k,j}^i)\}_{j=1}^{J}.$$
\textbf{Summarization.} 
For each reward candidate $n$ at iteration $i$, we aggregate the visual subtask scores across all $K$ trajectories and $J_i$ subtasks: $$\bar{s}_n^i = \mathbb{E}_{k,j}\!\left[s_{n,k,j}^i\right].$$
Additionally, for each subtask $j$, we aggregate the rationales $\{\rho_{n,k,j}^i\}_{k=1}^K$ into an analysis $p_{n,j}^i$ that identifies recurring failure modes and violations. The complete policy summary for candidate $n$ is:
$$G_n^i = (\bar{s}_n^i, \{p_{n,j}^i\}_{j=1}^{J}).$$
\textbf{Reward Selection.} 
We select the best-performing candidate at iteration $i$ by:
$$n^\star_i = \arg\max_{n \in \{1,\ldots,N\}} \bar{s}_n^i.$$
The reward $R_{n^\star_i}^i$, policy $\pi_{n^\star_i}^i$, and summary $G_{n^\star_i}^i$ are checkpointed as the current best solution.

\textbf{Subtask Reflection.}
Based on the summary $G_{n^\star_i}^i$ we prompt the VLM to analyze whether the current subtask decomposition $T^i=[t_j^i]_{j=1}^{J}$ sufficiently captures all aspects of the instruction $I$. 
The VLM proposes revisions $t^{i+1}_{j}$ for ambigious or underspecified subtasks and retains others ($t^{i+1}_{j} = t^i_j$), thereby conserving the total number of subtasks $J$.

\textbf{Reward Reflection.}
For each candidate $n$, we prompt the VLM with $(I, E, T^{i+1}, R_{n^{\star}_i}^i, G_{n^{\star}_i}^i)$ to generate an improved reward candidate. The VLM refines reward shaping, component weights, or introduces new reward terms to address the identified failure modes. This produces the next generation of rewards $\{R_n^{i+1}\}_{n=1}^N$.

\textbf{Termination.} After $M$ iterations, RDA returns the best reward found across all candidates.

\section{Experiments}
\label{sec:experiments}

We evaluate Reward Design Agent (RDA) on a diverse set of robotic control tasks to assess its ability to generate reward functions that achieve both instruction alignment and task success.

\subsection{Setup}

\begin{figure}[t]
\begin{center}
\includegraphics[width=.88\textwidth]{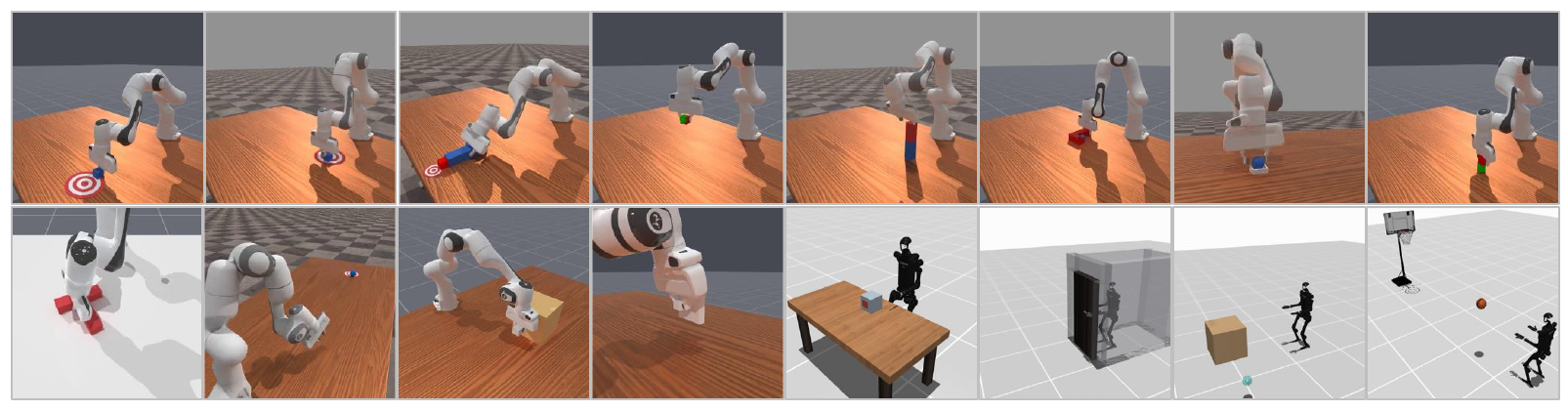}
\end{center}
\vspace{-3.5mm}
\caption{\textbf{Tasks.} 12 tabletop manipulation tasks from ManiSkill3~\citep{tao2024maniskill3} and the 4 whole-body manipulation tasks from HumanoidBench~\citep{sferrazza2024humanoidbench}.}
\vspace{-2mm}
\label{figure:tasks}
\end{figure}

\textbf{Tasks.}
As shown in Figure~\ref{figure:tasks}, our benchmark covers 16 tasks across two domains:
\begin{itemize}[leftmargin=10pt]
\item \textbf{ManiSkill~\citep{tao2024maniskill3}:} 12 tabletop manipulation tasks, such as pick-and-place, non-prehensile manipulation (e.g., rolling or dragging objects), and insertion tasks (e.g., inserting a peg or charger). We adopt the official task descriptions as the instructions for these tasks.
\item \textbf{HumanoidBench~\citep{sferrazza2024humanoidbench}:} 4 whole-body manipulation tasks with pushing objects, opening a door and exiting, moving a package to a target location, and performing a basketball layup. As the benchmark does not provide textual task descriptions, we author instructions based on the official task definitions to accurately capture the intended human objectives.
\end{itemize}

\textbf{Training.} RDA uses GPT-5~\citep{openai2025gpt5} with medium reasoning effort as the VLM backbone. Each run consists of $M=5$ evolutionary iterations. In each iteration, GPT-5 proposes multiple reward candidates, trains a policy under each candidate, and refines the best candidates based on visual trajectory analysis, summarization, and reflection.

All policies are trained using Soft Actor-Critic (SAC)~\citep{haarnoja2018soft} with the Simba architecture~\citep{lee2024simba, lee2025hyperspherical, kim2026flashsac}. Each task uses 3 random seeds. Hyperparameters are:
\begin{itemize}[leftmargin=10pt]
\item \textbf{ManiSkill:} 4 reward candidates per iteration, 50M env steps with 2048 environments.
\item \textbf{HumanoidBench:} 8 reward candidates per iteration, 10M env steps with 16 environments.
\end{itemize}

\textbf{Evaluation.}
We use two complementary metrics to assess policy:

\begin{itemize}[leftmargin=10pt]
\item \textbf{Alignment Rate:} Measures behavioral alignment with the task instruction that encodes human intention. For each trained policy, we collect 5 rollout videos and send each video with the task instruction to GPT-4.1~\citep{openai2025gpt4.1}. The VLM evaluates each video 4 times, rating instruction alignment on a 5-point Likert scale (1: complete misalignment, 5: perfect alignment). 
Unlike~\citet{muslimani2025towards}, we use point-wise rather than pairwise evaluation due to GPT's visual context limits. 
For each policy, we report the average score normalized in $[0,1]$.
\item \textbf{Success Rate:} Measures whether the policy achieves the task goal using the binary success indicator provided by each benchmark (e.g., object within target tolerance). We report the success rate over evaluation episodes. Unlike the alignment rate, this metric evaluates only task completion and does not assess whether the policy follows the intended behavior specified in the instruction.
\end{itemize}


\textbf{Baselines.} We compare RDA against three baselines:

\begin{itemize}[leftmargin=10pt]
\item \textbf{Human Sparse:} Binary success verifier as a sparse reward function (1 for success, 0 otherwise).
\item \textbf{Human Dense:} Hand-crafted dense reward functions provided in the original benchmarks.
\item \textbf{Eureka~\citep{ma2023eureka}:} The state-of-the-art LLM-based reward design method that uses evolutionary search to generate and refine reward code. The key difference from RDA is that Eureka reflects only on numerical reward statistics rather than on the visual analysis. We use the same VLM backbone, task instructions, training hyperparameters, and compute budget as RDA.
\end{itemize}

\begin{figure}[t!]
\begin{center}
\includegraphics[width=.99\textwidth, trim=5 5 5 5, clip]{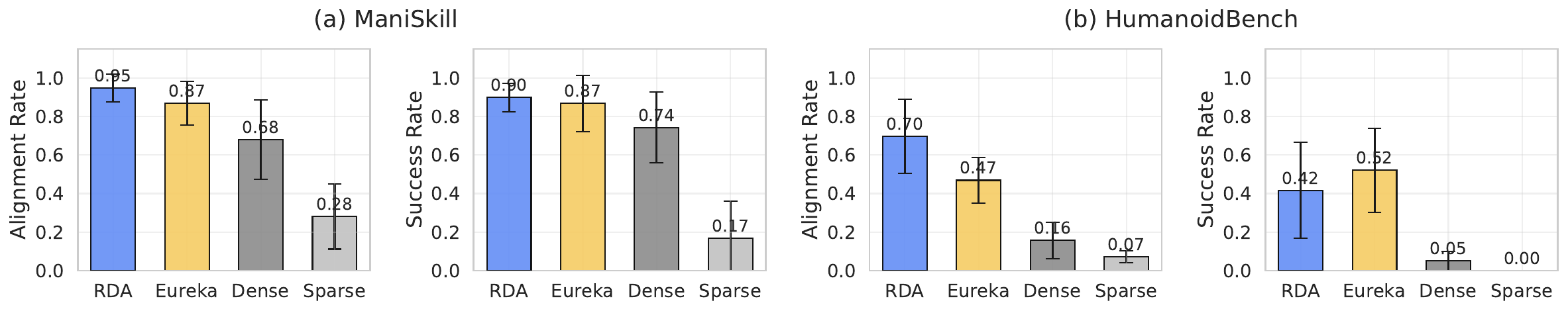}
\end{center}
\vspace{-3mm}
\caption{\textbf{Comparison of reward design methods on alignment and success rates.} 
\textbf{(a) ManiSkill:} Both RDA and Eureka substantially outperform human-designed rewards, achieving high alignment and success, demonstrating that LLM-based reward generation is effective for short-horizon tasks. \textbf{(b) HumanoidBench:} RDA achieves higher alignment rate and comparable success rate to Eureka, while human-designed rewards largely fail. The alignment gap shows that Eureka's lack of visual analysis causes reward mis-specification.}
\vspace{-1mm}
\label{figure:main_results}
\end{figure}

\subsection{Main Results}

Figure~\ref{figure:main_results} summarizes the performance of RDA and all baselines across both benchmarks. Full per-task results are provided in Appendix~\ref{appendix:full_result}.

\textbf{ManiSkill.} On the 12 tabletop manipulation tasks, RDA and Eureka substantially outperform human-designed rewards. RDA achieves the highest performance with 0.95 alignment and 0.90 success rate, marginally ahead of Eureka (0.87 alignment, 0.87 success). In contrast, human-designed dense rewards reach only 0.68 alignment and 0.74 success rate, while sparse rewards largely fail with 0.28 alignment and 0.17 success rate. Notably, RDA is the only method that solves PlugCharger, the most challenging task in the suite. The comparable performance of RDA and Eureka on these relatively short-horizon tasks suggests that for simpler manipulation scenarios, coarse reward statistics provide sufficient feedback for effective reward refinement.

\textbf{HumanoidBench.} On the 4 whole-body manipulation tasks, the advantage of visual feedback becomes pronounced. RDA achieves 0.70 alignment and 0.42 success rate, significantly outperforming Eureka in alignment (0.47, a 49\% relative improvement), while obtaining slightly lower success rate. This gap reveals a critical distinction: Eureka frequently solves tasks through misaligned behaviors—such as throwing the package to the goal rather than gently pushing it (Figure~\ref{figure:humanoid_package})—while RDA's visual trajectory analysis enables it to detect and correct such reward mis-specifications during the refinement loop. Human-designed rewards achieve only 0.25 alignment and 0.08 success, underscoring the difficulty of manually engineering rewards for complex, temporally extended tasks. These results demonstrate that as task complexity and temporal horizon increase, visual feedback becomes essential for maintaining instruction-alignment.


\begin{figure}[t!]
\centering
\footnotesize
\begin{tcolorbox}[enhanced, breakable=false, colback=pastel_light_blue!10, colframe=pastel_light_blue!70!black, title=(a) Instruction, left=2mm, top=1mm, bottom=0mm]
\begin{verbatim}
Walk towards the package, stand behind it, maintain contact, and push it to the goal location.
\end{verbatim}
\end{tcolorbox}

\footnotesize
\begin{tcolorbox}[breakable=false, colback=pastel_light_blue!10, colframe=pastel_light_blue!70!black, title=(b) Subtask, left=2mm, top=1mm, bottom=0mm]
\begin{verbatim}
3. Place both hands on the rear face of the box, establishing firm contact.
5. Push the box forward while maintaining balance and steady momentum.
\end{verbatim}
\end{tcolorbox}

\vspace{-4mm}
\caption*{\textbf{(c) Rolled-out Trajectory.}}
\label{figure:package_evolution}
\includegraphics[width=.99\textwidth]{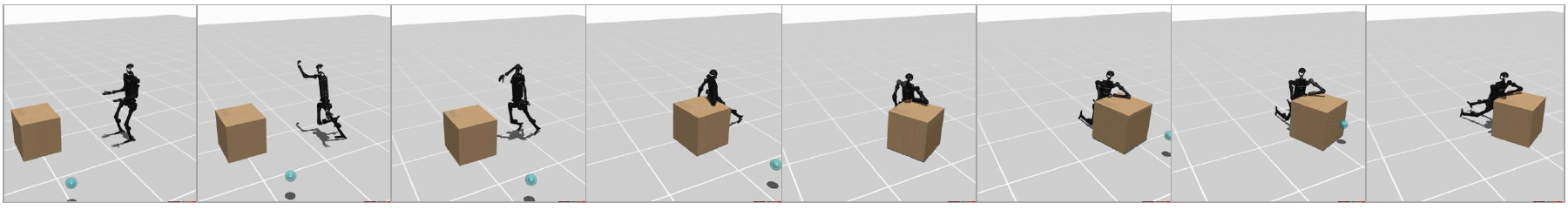}
\vspace{-1mm}

\begin{tcolorbox}[breakable=false, colback=pastel_light_blue!10, colframe=pastel_light_blue!70!black, 
title=(d) Visual Analysis, left=2mm, top=1mm, bottom=0mm]
\footnotesize
\begin{ColorVerbatim}
\textcolor{comment_green}{"subtask"}: 3
\textcolor{comment_green}{"score"}: 0.5,
\textcolor{comment_green}{"analysis"}: "The robot leans its upper body against the rear face of the box. 
  The hand-contact is present but weak (reward_hand_contact=0.14 at step 35, 0.25 at step 40), 
  indicating contact without firm or stable placement of both hands."
\end{ColorVerbatim}
\end{tcolorbox}

\footnotesize
\begin{tcolorbox}[breakable=false, colback=pastel_light_blue!10, colframe=pastel_light_blue!70!black, title=(e) Subtask Reflection, left=2mm, top=1mm, bottom=0mm]
\begin{ColorVerbatim}
\textcolor{comment_green}{"subtask"}: 3
\textcolor{comment_green}{"reasoning"}: 'Place both hands on the rear face of the box' was refined.
    \textcolor{comment_red}{The agent consistently pushed using its torso/forearms without firm bilateral hand contact.} 
    Reward signals for forward motion overshadowed the contact, leading to unnatural pushing. 
\textcolor{comment_green}{"revision"}: Place both hands on the box \textcolor{comment_red}{while maintaining  bilateral hand contact.}
\end{ColorVerbatim}
\end{tcolorbox}

\footnotesize
\begin{tcolorbox}[breakable=false, colback=pastel_light_blue!10, colframe=pastel_light_blue!70!black, title=(f) Reward Reflection, left=2mm, top=1mm, bottom=0mm]
\begin{ColorVerbatim}
def get_reward(self) -> Tuple[float, Dict]:
    \textcolor{comment_green}{# Subtask 3: Place both hands on rear face of the box (use proximity to rear plane)
    }
    \textcolor{comment_red}{# Modification: Bilateral contact gate, require both hands}
    reward_hand_contact_norm = float((left_contact + right_contact) / 2.0)
    \textcolor{comment_red}{bilateral_contact = float(min(left_contact, right_contact))}
    
    \textcolor{comment_green}{# Subtask 5: Push box forward while maintaining balance and modulate force for smooth motion}
    \textcolor{comment_red}{# Modification: gate forward motion by bilateral hand contact}
    forward_speed = float(np.dot(box_lin_vel, dir_hat)) 
    reward_speed_match_norm = exp_norm_sq(forward_speed - 0.40, scale=0.40)
    ang_speed = float(np.linalg.norm(box_ang_vel))
    reward_angular_stability_norm = exp_norm(ang_speed, scale=1.0)
    \textcolor{comment_red}{reward_box_forward = float(np.clip(reward_box_forward_raw * bilateral_contact, 0.0, 1.0))}

    return reward
\end{ColorVerbatim}
\end{tcolorbox}
\vspace{-3mm}
\caption{\textbf{RDA Evolution Process on Humanoid Package.}
In this example, the humanoid learns to push the package with its torso rather than both hands. RDA's visual analysis diagnoses this failure mode (weak bilateral hand contact) and refines the subtask specification and reward function to enforce pushing with bilateral contact.}
\vspace{-4mm}
\label{figure:package_response}
\end{figure}

\subsection{How does RDA progressively improve the reward function?}

To understand how RDA progressively improves reward quality, we analyze the evolution of subtasks and reward code through one refinement iteration on the humanoid package delivery task. Figure~\ref{figure:package_response} illustrates this process, focusing on two representative subtasks: \textit{establish hand contact} (subtask 3) and \textit{push box forward} (subtask 5). The complete refinement trace is in Appendix~\ref{appendix:rda_example}.

\textbf{Task Decomposition.} Given the natural language instruction \textit{"Walk towards the package, stand behind it, maintain contact, and push it to the goal location"}, RDA decomposes the task into seven interpretable subtasks. We focus on two critical subtasks: subtask 3 (\textit{establish hand contact with the box}) and subtask 5 (\textit{push the box forward to the target}).

\textbf{Trajectory Rollout.}
After training an RL policy with a candidate reward function, RDA collects rollout trajectories for evaluation. In this iteration, the humanoid fails to establish proper bilateral hand contact and instead leans its torso against the box to push it forward. While this strategy successfully moves the package toward the goal, it violates the intended pushing behavior specified in the task instruction.

\textbf{Visual Analysis.} RDA performs visual analysis of the rollout for each subtask. For subtask 3 (\textit{"Place both hands on the rear face of the box"}), the VLM observes: \textit{"The robot leans its upper body against the rear face of the box ... indicating contact without firm or stable placement of both hands."} This analysis provides fine-grained diagnostic information beyond binary success: it identifies not just that the subtask fails (assigned score: 0.5/1.0), but precisely \textit{how}—the agent achieves weak, asymmetric contact by leaning rather than using both hands properly.

\textbf{Subtask Reflection.} Guided by the visual diagnosis, RDA refines the subtask specification to make the intended behavior more explicit. 
Based on this visual analysis, RDA identifies the root cause, \textit{"The agent consistently pushed using its torso/forearms without firm bilateral hand contact"}, and revises the subtask description to: \textit{"Place both hands on the rear face of the box while maintaining continuous bilateral hand contact."} 

\textbf{Reward Reflection.} The refined subtask specification then informs concrete modifications to the reward code. For subtask 3, RDA introduces a \texttt{bilateral\_contact} term that requires both hands to be in contact with the box, and modifies the hand contact reward to depend on it: \texttt{reward\_hand\_contact = ... * bilateral\_contact}. For subtask 5, forward box motion is explicitly gated by bilateral contact: \texttt{reward\_box\_forward = ... * bilateral\_contact}, ensuring the agent cannot receive forward-progress reward unless it maintains proper hand placement.

\begin{figure}[h!]
\begin{minipage}{0.52\textwidth}
Figure~\ref{figure:rda_evolution} tracks how the refinement process progressively improves the reward function across iterations. 
The agent achieves full task success (1.0) after the second iteration, yet initial alignment is only 0.30; the box reaches the goal through leaning the torso. Through successive refinement cycles, alignment steadily improves to 0.68 by iteration 4, substantially surpassing both Eureka (0.28) and human-designed rewards (0.08).
This divergence between early success and gradual alignment underscores the core value of visual feedback: task completion alone is insufficient when the policy violates specifications. By iteration 4, RDA produces a policy that both succeeds and adheres to the intended instruction, as shown in Figure~\ref{figure:humanoid_package}.
\end{minipage}
\hspace{0.015\textwidth}
\begin{minipage}{0.46\textwidth}
\includegraphics[width=.99\textwidth, trim=6 6 6 6, clip]{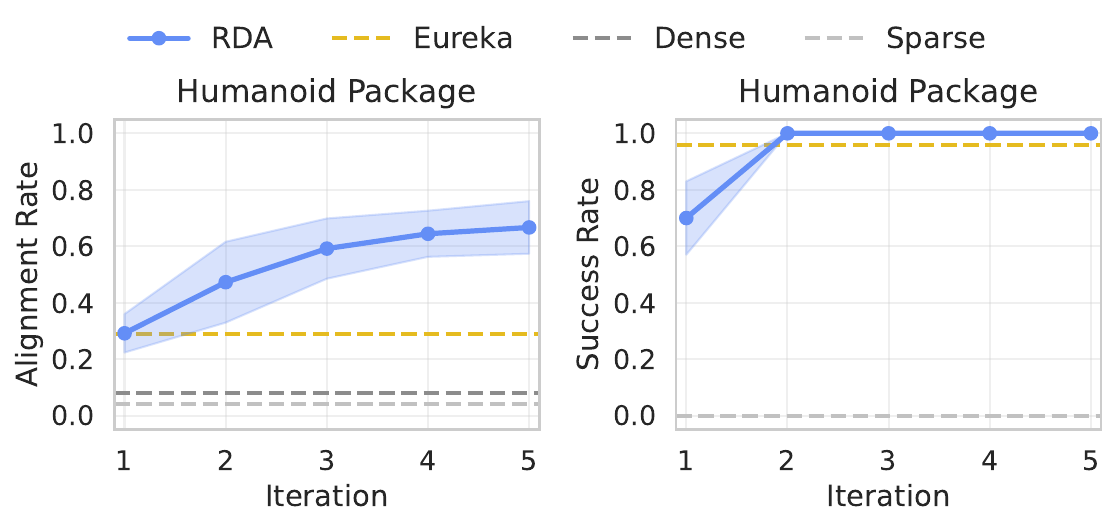}
    \vspace{-5mm}
    \caption{\textbf{Evolution of the RDA process.} Alignment and success rate over RDA iterations on the humanoid package task.RDA achieves full task success after the first iteration but initially exhibits low alignment (0.30). Visual feedback progressively improves alignment to 0.68 at the fifth iteration,  outperforming baseline methods.}
    \vspace{-1mm}
    \label{figure:rda_evolution}
\end{minipage}
\end{figure}

\subsection{Ablation Study}

To understand which components drive RDA's performance and how computational budget affects results, we conduct two ablation studies on representative tasks: PlugCharger (ManiSkill) and Package (HumanoidBench). Results are averaged over 3 random seeds.

\begin{figure}[t]
\begin{center}
\includegraphics[width=.99\textwidth]{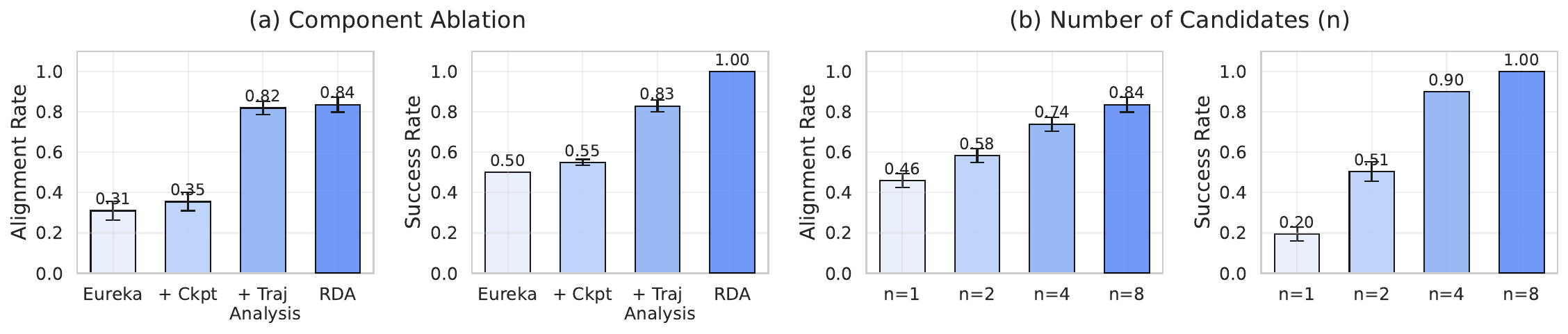}
\end{center}
\vspace{-4mm}
\caption{\textbf{Ablation Studies.} \textbf{(a) Component ablation:} Progressive component addition reveals that visual trajectory analysis is critical for alignment, while subtask decomposition (Traj Analysis → RDA) further improves success through better credit assignment.
\textbf{(b) Scaling analysis:} Performance improves with more reward candidates per iteration.}
\vspace{-1mm}
\label{figure:ablation}
\end{figure}

\textbf{Component Analysis.} Figure~\ref{figure:ablation}.(a) progressively adds components to the Eureka baseline. Loading the best checkpoint from the previous iteration (+Ckpt) provides modest gains (alignment: 0.31 → 0.35; success: 0.50 → 0.55), indicating that warm-starting aids exploration but remains limited without better reflection signals.

The most substantial gain comes from trajectory-based visual analysis (+Traj Analysis), which boosts alignment to 0.82 and success to 0.83 without subtask decomposition. This result demonstrates that visual feedback is the critical mechanism for detecting and correcting misalignment.

The full \textbf{RDA} system adds subtask decomposition for both reward generation and analysis, achieving 0.84 alignment and 1.0 success rate. The subtask structure primarily improves success on PlugCharger (0.83 → 1.0), which we attribute to better credit assignment: explicitly decomposing the task helps assign credit to the fine-grained manipulation of inserting the charger into the small receptacle.

\textbf{Scaling Analysis.} Figure~\ref{figure:ablation}.(b) varies the number of candidate reward functions generated per iteration. Performance scales consistently with compute: increasing from 1 to 8 candidates improves alignment from 0.46 to 0.84 and success from 0.20 to 1.0. Notably, even 2 candidates yield substantial gains over 
1 candidate (alignment: 0.46 → 0.58; success: 0.20 → 0.51), indicating that RDA's evolutionary search benefits from larger candidate pools that provide more diverse starting points.

\section{Conclusion}

We introduced RDA, a VLM-based agentic framework that automates reward function design through task decomposition, visual trajectory analysis, and iterative refinement. By closing the loop between policy learning and visual diagnostics, RDA addresses a critical gap in existing automated reward design methods: the inability to detect and correct misaligned behaviors.

\textbf{Lessons.} Our experiments on ManiSkill and HumanoidBench confirm this capability. RDA matches or exceeds the success rate of human-designed and Eureka-generated rewards while substantially improving instruction alignment (Figure~\ref{figure:main_results}). Ablations reveal that visual trajectory analysis drives alignment improvement, while subtask decomposition enhances success through better credit assignment (Figure~\ref{figure:ablation}).

\textbf{Limitations and Future Work.} We point out certain limitations of RDA and leave their exploration for future work: (1) RDA performs better with more reward candidates, but each candidate requires a full round of RL training, which is costly. Internal reflection mechanisms~\citep{lu2024ai} could self-diagnose proposed reward functions even before RL training, thereby reducing wasted computation. (2) RDA's diagnostic capabilities are constrained by the VLM's limited context length, high inference cost, and imperfect fine-grained visual reasoning. However, these capabilities will continue to improve as VLMs improve. (3) RDA struggles when earlier subtask rewards conflict with those of later subtasks. For example, in the Humanoid Basketball task, we consistently observed that catching rewards suppresses the exploration for throwing the basketball into the goal. Conditioning the policy on explicit subtask states could enable better exploration for all subtasks.

Despite these limitations, RDA demonstrates that vision-guided agentic loops significantly improve both alignment and success rate in automated reward design. As foundation models and simulators scale, we believe that a vision-based agentic framework will become essential for learning complex, long-horizon tasks with minimal human supervision.

\newpage


\bibliography{main}

\begin{thebibliography}{57}
\providecommand{\natexlab}[1]{#1}
\providecommand{\url}[1]{\texttt{#1}}
\expandafter\ifx\csname urlstyle\endcsname\relax
  \providecommand{\doi}[1]{DOI: #1}\else
  \providecommand{\doi}{DOI: \begingroup \urlstyle{rm}\Url}\fi

\bibitem[Abbeel \& Ng(2004)Abbeel and Ng]{abbeel2004apprenticeship}
Pieter Abbeel and Andrew~Y Ng.
\newblock Apprenticeship learning via inverse reinforcement learning.
\newblock In \emph{Proc. the International Conference on Machine Learning (ICML)}, pp.\ ~1, 2004.

\bibitem[Andrychowicz et~al.(2020)Andrychowicz, Baker, Chociej, Jozefowicz, McGrew, Pachocki, Petron, Plappert, Powell, Ray, et~al.]{andrychowicz2020learning}
OpenAI:~Marcin Andrychowicz, Bowen Baker, Maciek Chociej, Rafal Jozefowicz, Bob McGrew, Jakub Pachocki, Arthur Petron, Matthias Plappert, Glenn Powell, Alex Ray, et~al.
\newblock Learning dexterous in-hand manipulation.
\newblock \emph{The International Journal of Robotics Research}, 39\penalty0 (1):\penalty0 3--20, 2020.

\bibitem[Baumli et~al.(2023)Baumli, Baveja, Behbahani, Chan, Comanici, Flennerhag, Gazeau, Holsheimer, Horgan, Laskin, et~al.]{baumli2023vlmsource}
Kate Baumli, Satinder Baveja, Feryal Behbahani, Harris Chan, Gheorghe Comanici, Sebastian Flennerhag, Maxime Gazeau, Kristian Holsheimer, Dan Horgan, Michael Laskin, et~al.
\newblock Vision-language models as a source of rewards.
\newblock \emph{arXiv preprint arXiv:2312.09187}, 2023.

\bibitem[Booth et~al.(2023)Booth, Knox, Shah, Niekum, Stone, and Allievi]{booth2023perils}
Serena Booth, W~Bradley Knox, Julie Shah, Scott Niekum, Peter Stone, and Alessandro Allievi.
\newblock The perils of trial-and-error reward design: misdesign through overfitting and invalid task specifications.
\newblock In \emph{Proceedings of the AAAI Conference on Artificial Intelligence}, volume~37, pp.\  5920--5929, 2023.

\bibitem[Chiang et~al.(2019)Chiang, Faust, Fiser, and Francis]{chiang2019learning}
Hao-Tien~Lewis Chiang, Aleksandra Faust, Marek Fiser, and Anthony Francis.
\newblock Learning navigation behaviors end-to-end with autorl.
\newblock \emph{IEEE Robotics and Automation Letters}, 4\penalty0 (2):\penalty0 2007--2014, 2019.

\bibitem[Christiano et~al.(2017)Christiano, Leike, Brown, Martic, Legg, and Amodei]{christiano2017deep}
Paul~F Christiano, Jan Leike, Tom Brown, Miljan Martic, Shane Legg, and Dario Amodei.
\newblock Deep reinforcement learning from human preferences.
\newblock \emph{Advances in neural information processing systems}, 30, 2017.

\bibitem[Chung et~al.(2026)Chung, Ha, Kwak, Kwon, Lee, Lee, and Lee]{prompt2policy2026}
Wooseong Chung, Taegwan Ha, Yunhyeok Kwak, Taehwan Kwon, Jeong-Gwan Lee, Kangwook Lee, and Suyoung Lee.
\newblock Prompt-to-policy: Agentic engineering for reinforcement learning, 2026.
\newblock URL \url{https://github.com/krafton-ai/Prompt2Policy}.

\bibitem[Cui et~al.(2025)Cui, Liu, Meng, Yu, Song, Zhang, Zhu, and Huang]{cui2025grove}
Jieming Cui, Tengyu Liu, Ziyu Meng, Jiale Yu, Ran Song, Wei Zhang, Yixin Zhu, and Siyuan Huang.
\newblock Grove: A generalized reward for learning open-vocabulary physical skill.
\newblock In \emph{Proceedings of the Computer Vision and Pattern Recognition Conference}, pp.\  15781--15790, 2025.

\bibitem[Escontrela et~al.(2023)Escontrela, Adeniji, Yan, Jain, Peng, Goldberg, Lee, Hafner, and Abbeel]{escontrela2023viper}
Alejandro Escontrela, Ademi Adeniji, Wilson Yan, Ajay Jain, Xue~Bin Peng, Ken Goldberg, Youngwoon Lee, Danijar Hafner, and Pieter Abbeel.
\newblock Video prediction models as rewards for reinforcement learning.
\newblock \emph{Advances in Neural Information Processing Systems}, 36:\penalty0 68760--68783, 2023.

\bibitem[Faust et~al.(2019)Faust, Francis, and Mehta]{faust2019evolving}
Aleksandra Faust, Anthony Francis, and Dar Mehta.
\newblock Evolving rewards to automate reinforcement learning.
\newblock \emph{arXiv preprint arXiv:1905.07628}, 2019.

\bibitem[Haarnoja et~al.(2018)Haarnoja, Zhou, Hartikainen, Tucker, Ha, Tan, Kumar, Zhu, Gupta, Abbeel, et~al.]{haarnoja2018soft}
Tuomas Haarnoja, Aurick Zhou, Kristian Hartikainen, George Tucker, Sehoon Ha, Jie Tan, Vikash Kumar, Henry Zhu, Abhishek Gupta, Pieter Abbeel, et~al.
\newblock Soft actor-critic algorithms and applications.
\newblock \emph{arXiv preprint arXiv:1812.05905}, 2018.

\bibitem[Hejna~III \& Sadigh(2023)Hejna~III and Sadigh]{hejna2023few}
Donald~Joseph Hejna~III and Dorsa Sadigh.
\newblock Few-shot preference learning for human-in-the-loop rl.
\newblock In \emph{Conference on Robot Learning}, pp.\  2014--2025. PMLR, 2023.

\bibitem[Ho \& Ermon(2016)Ho and Ermon]{ho2016gail}
Jonathan Ho and Stefano Ermon.
\newblock Generative adversarial imitation learning.
\newblock \emph{Advances in neural information processing systems}, 29, 2016.

\bibitem[Hu \& Sadigh(2023)Hu and Sadigh]{hu2023language}
Hengyuan Hu and Dorsa Sadigh.
\newblock Language instructed reinforcement learning for human-ai coordination.
\newblock In \emph{International Conference on Machine Learning}, pp.\  13584--13598. PMLR, 2023.

\bibitem[Hung et~al.(2025)Hung, Lo, Yeh, Hsu, Chen, and Hsu]{hung2025victor}
Kuo-Han Hung, Pang-Chi Lo, Jia-Fong Yeh, Han-Yuan Hsu, Yi-Ting Chen, and Winston~H. Hsu.
\newblock Victor: Learning hierarchical vision-instruction correlation rewards for long-horizon manipulation.
\newblock In \emph{The Thirteenth International Conference on Learning Representations (ICLR)}, 2025.

\bibitem[Hwangbo et~al.(2019)Hwangbo, Lee, Dosovitskiy, Bellicoso, Tsounis, Koltun, and Hutter]{hwangbo2019learning}
Jemin Hwangbo, Joonho Lee, Alexey Dosovitskiy, Dario Bellicoso, Vassilios Tsounis, Vladlen Koltun, and Marco Hutter.
\newblock Learning agile and dynamic motor skills for legged robots.
\newblock \emph{Science Robotics}, 4\penalty0 (26):\penalty0 eaau5872, 2019.

\bibitem[Kim et~al.(2026)Kim, Lee, Park, Kim, Nahendra, Seno, Min, Palenicek, Vogt, Kragic, et~al.]{kim2026flashsac}
Donghu Kim, Youngdo Lee, Minho Park, Kinam Kim, I~Nahendra, Takuma Seno, Sehee Min, Daniel Palenicek, Florian Vogt, Danica Kragic, et~al.
\newblock Flashsac: Fast and stable off-policy reinforcement learning for high-dimensional robot control.
\newblock \emph{arXiv preprint arXiv:2604.04539}, 2026.

\bibitem[Knox et~al.(2023)Knox, Allievi, Banzhaf, Schmitt, and Stone]{knox2023reward}
W~Bradley Knox, Alessandro Allievi, Holger Banzhaf, Felix Schmitt, and Peter Stone.
\newblock Reward (mis) design for autonomous driving.
\newblock \emph{Artificial Intelligence}, 316:\penalty0 103829, 2023.

\bibitem[Kostrikov et~al.(2018)Kostrikov, Agrawal, Dwibedi, Levine, and Tompson]{kostrikov2018discriminator}
Ilya Kostrikov, Kumar~Krishna Agrawal, Debidatta Dwibedi, Sergey Levine, and Jonathan Tompson.
\newblock Discriminator-actor-critic: Addressing sample inefficiency and reward bias in adversarial imitation learning.
\newblock \emph{arXiv preprint arXiv:1809.02925}, 2018.

\bibitem[Kwon et~al.(2023)Kwon, Xie, Bullard, and Sadigh]{kwon2023rdllm}
Minae Kwon, Sang~Michael Xie, Kalesha Bullard, and Dorsa Sadigh.
\newblock Reward design with language models.
\newblock \emph{arXiv preprint arXiv:2303.00001}, 2023.

\bibitem[Lee et~al.(2024)Lee, Hwang, Kim, Kim, Tai, Subramanian, Wurman, Choo, Stone, and Seno]{lee2024simba}
Hojoon Lee, Dongyoon Hwang, Donghu Kim, Hyunseung Kim, Jun~Jet Tai, Kaushik Subramanian, Peter~R Wurman, Jaegul Choo, Peter Stone, and Takuma Seno.
\newblock Simba: Simplicity bias for scaling up parameters in deep reinforcement learning.
\newblock \emph{arXiv preprint arXiv:2410.09754}, 2024.

\bibitem[Lee et~al.(2025)Lee, Lee, Seno, Kim, Stone, and Choo]{lee2025hyperspherical}
Hojoon Lee, Youngdo Lee, Takuma Seno, Donghu Kim, Peter Stone, and Jaegul Choo.
\newblock Hyperspherical normalization for scalable deep reinforcement learning.
\newblock \emph{arXiv preprint arXiv:2502.15280}, 2025.

\bibitem[Lee et~al.(2021)Lee, Smith, and Abbeel]{lee2021pebble}
Kimin Lee, Laura~M Smith, and Pieter Abbeel.
\newblock Pebble: Feedback-efficient interactive reinforcement learning via relabeling experience and unsupervised pre-training.
\newblock In \emph{Proc. the International Conference on Machine Learning (ICML)}, pp.\  6152--6163. PMLR, 2021.

\bibitem[Lee et~al.(2026)Lee, Wagenmaker, Pertsch, Liang, Levine, and Finn]{lee2026roboreward}
Tony Lee, Andrew Wagenmaker, Karl Pertsch, Percy Liang, Sergey Levine, and Chelsea Finn.
\newblock Roboreward: General-purpose vision-language reward models for robotics.
\newblock \emph{arXiv preprint arXiv:2601.00675}, 2026.

\bibitem[Lu et~al.(2024)Lu, Lu, Lange, Foerster, Clune, and Ha]{lu2024ai}
Chris Lu, Cong Lu, Robert~Tjarko Lange, Jakob Foerster, Jeff Clune, and David Ha.
\newblock The ai scientist: Towards fully automated open-ended scientific discovery.
\newblock \emph{arXiv preprint arXiv:2408.06292}, 2024.

\bibitem[Luu et~al.(2025{\natexlab{a}})Luu, Lee, Lee, and Yoo]{luu2025policy}
Tung~M Luu, Donghoon Lee, Younghwan Lee, and Chang~D Yoo.
\newblock Policy learning from large vision-language model feedback without reward modeling.
\newblock \emph{arXiv preprint arXiv:2507.23391}, 2025{\natexlab{a}}.

\bibitem[Luu et~al.(2025{\natexlab{b}})Luu, Lee, Lee, Kim, Kim, and Yoo]{luu2025erlvlm}
Tung~Minh Luu, Younghwan Lee, Donghoon Lee, Sunho Kim, Min~Jun Kim, and Chang~D Yoo.
\newblock Enhancing rating-based reinforcement learning to effectively leverage feedback from large vision-language models.
\newblock \emph{arXiv preprint arXiv:2506.12822}, 2025{\natexlab{b}}.

\bibitem[Ma et~al.(2024)Ma, Liang, Wang, Zhu, Fan, Bastani, and Jayaraman]{ma2024dreureka}
Jason Ma, William Liang, Hung-Ju Wang, Yuke Zhu, Linxi Fan, Osbert Bastani, and Dinesh Jayaraman.
\newblock Dreureka: Language model guided sim-to-real transfer.
\newblock \emph{RSS}, 2024.

\bibitem[Ma et~al.(2025{\natexlab{a}})Ma, Seno, Subramanian, Wurman, Stone, and Sherstan]{ma2025automated}
Michel Ma, Takuma Seno, Kaushik Subramanian, Peter~R Wurman, Peter Stone, and Craig Sherstan.
\newblock Automated reward design for gran turismo.
\newblock \emph{arXiv preprint arXiv:2511.02094}, 2025{\natexlab{a}}.

\bibitem[Ma et~al.(2022)Ma, Sodhani, Jayaraman, Bastani, Kumar, and Zhang]{ma2022vip}
Yecheng~Jason Ma, Shagun Sodhani, Dinesh Jayaraman, Osbert Bastani, Vikash Kumar, and Amy Zhang.
\newblock Vip: Towards universal visual reward and representation via value-implicit pre-training.
\newblock \emph{arXiv preprint arXiv:2210.00030}, 2022.

\bibitem[Ma et~al.(2023)Ma, Liang, Wang, Huang, Bastani, Jayaraman, Zhu, Fan, and Anandkumar]{ma2023eureka}
Yecheng~Jason Ma, William Liang, Guanzhi Wang, De-An Huang, Osbert Bastani, Dinesh Jayaraman, Yuke Zhu, Linxi Fan, and Anima Anandkumar.
\newblock Eureka: Human-level reward design via coding large language models.
\newblock \emph{arXiv preprint arXiv:2310.12931}, 2023.

\bibitem[Ma et~al.(2025{\natexlab{b}})Ma, Hejna, Fu, Shah, Liang, Xu, Kirmani, Xu, Driess, Xiao, et~al.]{ma2024gvl}
Yecheng~Jason Ma, Joey Hejna, Chuyuan Fu, Dhruv Shah, Jacky Liang, Zhuo Xu, Sean Kirmani, Peng Xu, Danny Driess, Ted Xiao, et~al.
\newblock Vision language models are in-context value learners.
\newblock In \emph{Proc. the International Conference on Learning Representations (ICLR)}, 2025{\natexlab{b}}.

\bibitem[Makoviychuk et~al.(2021)Makoviychuk, Wawrzyniak, Guo, Lu, Storey, Macklin, Hoeller, Rudin, Allshire, Handa, et~al.]{makoviychuk2021isaac}
Viktor Makoviychuk, Lukasz Wawrzyniak, Yunrong Guo, Michelle Lu, Kier Storey, Miles Macklin, David Hoeller, Nikita Rudin, Arthur Allshire, Ankur Handa, et~al.
\newblock Isaac gym: High performance gpu-based physics simulation for robot learning.
\newblock \emph{arXiv preprint arXiv:2108.10470}, 2021.

\bibitem[Muslimani et~al.(2025)Muslimani, Johnstonbaugh, Chandramouli, Booth, Knox, and Taylor]{muslimani2025towards}
Calarina Muslimani, Kerrick Johnstonbaugh, Suyog Chandramouli, Serena Booth, W~Bradley Knox, and Matthew~E Taylor.
\newblock Towards improving reward design in rl: A reward alignment metric for rl practitioners.
\newblock \emph{arXiv preprint arXiv:2503.05996}, 2025.

\bibitem[Myers et~al.(2022)Myers, Biyik, Anari, and Sadigh]{myers2022learning}
Vivek Myers, Erdem Biyik, Nima Anari, and Dorsa Sadigh.
\newblock Learning multimodal rewards from rankings.
\newblock In \emph{Conference on robot learning}, pp.\  342--352. PMLR, 2022.

\bibitem[Nahrendra et~al.(2026)Nahrendra, Lee, Lee, and Myung]{nahrendra2026locovlm}
I~Nahrendra, Seunghyun Lee, Dongkyu Lee, and Hyun Myung.
\newblock Locovlm: Grounding vision and language for adapting versatile legged locomotion policies.
\newblock \emph{arXiv preprint arXiv:2602.10399}, 2026.

\bibitem[Ng et~al.(2000)Ng, Russell, et~al.]{ng2000algorithms}
Andrew~Y Ng, Stuart Russell, et~al.
\newblock Algorithms for inverse reinforcement learning.
\newblock In \emph{Proc. the International Conference on Machine Learning (ICML)}, volume~1, pp.\ ~2, 2000.

\bibitem[Niekum et~al.(2010)Niekum, Barto, and Spector]{niekum2010genetic}
Scott Niekum, Andrew~G Barto, and Lee Spector.
\newblock Genetic programming for reward function search.
\newblock \emph{IEEE Transactions on Autonomous Mental Development}, 2\penalty0 (2):\penalty0 83--90, 2010.

\bibitem[OpenAI(2025{\natexlab{a}})]{openai2025gpt4.1}
OpenAI.
\newblock Gpt-4.1, 2025{\natexlab{a}}.
\newblock URL \url{https://openai.com/index/gpt-4-1/}.

\bibitem[OpenAI(2025{\natexlab{b}})]{openai2025gpt5}
OpenAI.
\newblock Gpt-5, 2025{\natexlab{b}}.
\newblock URL \url{https://openai.com/index/introducing-gpt-5/}.

\bibitem[Peng et~al.(2018)Peng, Kanazawa, Toyer, Abbeel, and Levine]{peng2018variational}
Xue~Bin Peng, Angjoo Kanazawa, Sam Toyer, Pieter Abbeel, and Sergey Levine.
\newblock Variational discriminator bottleneck: Improving imitation learning, inverse rl, and gans by constraining information flow.
\newblock \emph{arXiv preprint arXiv:1810.00821}, 2018.

\bibitem[Peng et~al.(2021)Peng, Ma, Abbeel, Levine, and Kanazawa]{peng2021amp}
Xue~Bin Peng, Ze~Ma, Pieter Abbeel, Sergey Levine, and Angjoo Kanazawa.
\newblock Amp: Adversarial motion priors for stylized physics-based character control.
\newblock \emph{ACM Transactions on Graphics (ToG)}, 40\penalty0 (4):\penalty0 1--20, 2021.

\bibitem[Rocamonde et~al.(2023)Rocamonde, Montesinos, Nava, Perez, and Lindner]{rocamonde2023vlmrl}
Juan Rocamonde, Victoriano Montesinos, Elvis Nava, Ethan Perez, and David Lindner.
\newblock Vision-language models are zero-shot reward models for reinforcement learning.
\newblock \emph{arXiv preprint arXiv:2310.12921}, 2023.

\bibitem[Sferrazza et~al.(2024)Sferrazza, Huang, Lin, Lee, and Abbeel]{sferrazza2024humanoidbench}
Carmelo Sferrazza, Dun-Ming Huang, Xingyu Lin, Youngwoon Lee, and Pieter Abbeel.
\newblock Humanoidbench: Simulated humanoid benchmark for whole-body locomotion and manipulation.
\newblock \emph{arXiv preprint arXiv:2403.10506}, 2024.

\bibitem[Singh et~al.(2009)Singh, Lewis, and Barto]{singh2009rewards}
Satinder Singh, Richard~L Lewis, and Andrew~G Barto.
\newblock Where do rewards come from.
\newblock In \emph{Proceedings of the annual conference of the cognitive science society}, pp.\  2601--2606. Cognitive Science Society, 2009.

\bibitem[Sontakke et~al.(2023)Sontakke, Zhang, Arnold, Pertsch, B{\i}y{\i}k, Sadigh, Finn, and Itti]{sontakke2023roboclip}
Sumedh Sontakke, Jesse Zhang, S{\'e}b Arnold, Karl Pertsch, Erdem B{\i}y{\i}k, Dorsa Sadigh, Chelsea Finn, and Laurent Itti.
\newblock Roboclip: One demonstration is enough to learn robot policies.
\newblock \emph{Advances in Neural Information Processing Systems}, 36:\penalty0 55681--55693, 2023.

\bibitem[Tang et~al.(2025)Tang, Abbatematteo, Hu, Chandra, Mart{\'\i}n-Mart{\'\i}n, and Stone]{tang8deep}
Chen Tang, Ben Abbatematteo, Jiaheng Hu, Rohan Chandra, Roberto Mart{\'\i}n-Mart{\'\i}n, and Peter Stone.
\newblock Deep reinforcement learning for robotics: A survey of real-world successes.
\newblock \emph{Annual Review of Control, Robotics, and Autonomous Systems}, 8, 2025.

\bibitem[Tao et~al.(2024)Tao, Xiang, Shukla, Qin, Hinrichsen, Yuan, Bao, Lin, Liu, Chan, et~al.]{tao2024maniskill3}
Stone Tao, Fanbo Xiang, Arth Shukla, Yuzhe Qin, Xander Hinrichsen, Xiaodi Yuan, Chen Bao, Xinsong Lin, Yulin Liu, Tse-kai Chan, et~al.
\newblock Maniskill3: Gpu parallelized robotics simulation and rendering for generalizable embodied ai.
\newblock \emph{arXiv preprint arXiv:2410.00425}, 2024.

\bibitem[Tessler et~al.(2024)Tessler, Guo, Nabati, Chechik, and Peng]{tessler2024maskedmimic}
Chen Tessler, Yunrong Guo, Ofir Nabati, Gal Chechik, and Xue~Bin Peng.
\newblock Maskedmimic: Unified physics-based character control through masked motion inpainting.
\newblock \emph{ACM Transactions on Graphics (TOG)}, 43\penalty0 (6):\penalty0 1--21, 2024.

\bibitem[Todorov et~al.(2012)Todorov, Erez, and Tassa]{todorov2012mujoco}
Emanuel Todorov, Tom Erez, and Yuval Tassa.
\newblock Mujoco: A physics engine for model-based control.
\newblock In \emph{2012 IEEE/RSJ international conference on intelligent robots and systems}, pp.\  5026--5033. IEEE, 2012.

\bibitem[Wang et~al.(2024)Wang, Sun, Zhang, Xian, Biyik, Held, and Erickson]{wang2024rl_vlmf}
Yufei Wang, Zhanyi Sun, Jesse Zhang, Zhou Xian, Erdem Biyik, David Held, and Zackory Erickson.
\newblock Rl-vlm-f: Reinforcement learning from vision language foundation model feedback.
\newblock \emph{arXiv preprint arXiv:2402.03681}, 2024.

\bibitem[Wilde et~al.(2022)Wilde, Biyik, Sadigh, and Smith]{wilde2022learning}
Nils Wilde, Erdem Biyik, Dorsa Sadigh, and Stephen~L Smith.
\newblock Learning reward functions from scale feedback.
\newblock In \emph{Conference on Robot Learning}, pp.\  353--362. PMLR, 2022.

\bibitem[Yang et~al.(2024)Yang, Jun, Tien, Russell, Dragan, and B{\i}y{\i}k]{yang2024trajectory}
Zhaojing Yang, Miru Jun, Jeremy Tien, Stuart~J Russell, Anca Dragan, and Erdem B{\i}y{\i}k.
\newblock Trajectory improvement and reward learning from comparative language feedback.
\newblock \emph{arXiv preprint arXiv:2410.06401}, 2024.

\bibitem[Yu et~al.(2023)Yu, Gileadi, Fu, Kirmani, Lee, Arenas, Chiang, Erez, Hasenclever, Humplik, et~al.]{yu2023language}
Wenhao Yu, Nimrod Gileadi, Chuyuan Fu, Sean Kirmani, Kuang-Huei Lee, Montse~Gonzalez Arenas, Hao-Tien~Lewis Chiang, Tom Erez, Leonard Hasenclever, Jan Humplik, et~al.
\newblock Language to rewards for robotic skill synthesis.
\newblock \emph{arXiv preprint arXiv:2306.08647}, 2023.

\bibitem[Zhai et~al.(2025)Zhai, Zhang, Zhang, Huang, Zhang, Zhou, Zhang, Liu, Lin, and Pang]{zhai2025vlac}
Shaopeng Zhai, Qi~Zhang, Tianyi Zhang, Fuxian Huang, Haoran Zhang, Ming Zhou, Shengzhe Zhang, Litao Liu, Sixu Lin, and Jiangmiao Pang.
\newblock A vision-language-action-critic model for robotic real-world reinforcement learning.
\newblock \emph{arXiv preprint arXiv:2509.15937}, 2025.

\bibitem[Zhang et~al.(2025)Zhang, Luo, Anwar, Sontakke, Lim, Thomason, Biyik, and Zhang]{zhang2025rewind}
Jiahui Zhang, Yusen Luo, Abrar Anwar, Sumedh~Anand Sontakke, Joseph~J Lim, Jesse Thomason, Erdem Biyik, and Jesse Zhang.
\newblock Rewind: Language-guided rewards teach robot policies without new demonstrations.
\newblock \emph{arXiv preprint arXiv:2505.10911}, 2025.

\bibitem[Ziebart et~al.(2008)Ziebart, Maas, Bagnell, Dey, et~al.]{ziebart2008maximum}
Brian~D Ziebart, Andrew~L Maas, J~Andrew Bagnell, Anind~K Dey, et~al.
\newblock Maximum entropy inverse reinforcement learning.
\newblock In \emph{Aaai}, volume~8, pp.\  1433--1438. Chicago, IL, USA, 2008.

\end{thebibliography}
\bibliographystyle{rlj}

\beginSupplementaryMaterials


\section{RDA Prompts}
\label{appendix:rda_prompt}

\subsection{Subtask Generation}

\begin{tcolorbox}[breakable, colback=pastel_light_blue!10, colframe=pastel_light_blue!70!black, title=Subtask Generation System Prompt]
\small
\begin{ColorVerbatim}
You are a task decomposition expert assisting reinforcement learning agents.
Your job is to break a given task into concise, 
high-level subtasks that describe the main stages required for success.

\textcolor{comment_green}{## Instructions}
1. Understand the task
Use the following inputs to infer how the agent should behave:
- Task instruction: outlines the goal and intended agent behavior.
- Environment code: includes the agent's observation and **success condition**.
- Image: visualizes the agent and environment.

2. Decompose the task into subtasks
Create a sequence of goal-oriented subtasks that describe progress toward completion.
- Each subtask should represent a distinct, measurable phase of progress.
- Do not include perception or reasoning steps (e.g., “identify goal”, “plan path”).
- Avoid numeric thresholds unless they are explicitly defined in success condition.
- Keep the subtasks concise.
- Use the fewest subtasks needed to describe essential behavior:
    - Simple tasks: ~5 subtasks
    - Complex or long-horizon tasks: ~10 subtasks
- The final subtask describes the agent achieving the success as an accomplished goal.
    - Do not phrase it as stop/halt.
    - Express the agent actively bringing about the success within the threshold 
    (e.g., “place the object within 0.05 m”).

\textcolor{comment_green}{## Output}
- Return a numbered list (1., 2., …) of subtasks in order.
\end{ColorVerbatim}
\end{tcolorbox}

\begin{tcolorbox}[breakable, colback=pastel_light_blue!10, colframe=pastel_light_blue!70!black, title=Subtask Generation User Prompt]
\small
\begin{verbatim}
You are given the following input to decompose the task into high-level subtasks:
- Task instruction: outlines the goal and intended agent behavior.
- Environment code: includes agent's observation and **success condition**.
- Image: visualizes the agent and environment.

Using this information, produce a numbered list of concise, logically ordered subtasks 
that describe the main stages of progress required for the agent to achieve success.

## Task Instruction
{task_instruction}

## Environment Code
{env_code}

## Image
{image}
\end{verbatim}
\end{tcolorbox}
\newpage

\subsection{Reward Generation}

\begin{tcolorbox}[breakable, colback=pastel_light_blue!10, colframe=pastel_light_blue!70!black, title=Reward Generation System Prompt]
\small
\begin{verbatim}
You are a reward engineering expert assisting reinforcement learning agents.
Your job is to design a well-shaped reward that maximizes the likelihood of task success.

## Instructions
1. Understand the task
Use the following inputs to understand the task:
- task instruction: outlines the goal and intended agent behavior.
- Subtask list: defines sequential phases toward success.
- Environment code: includes the agent's observation and **success condition**.

2. Implement the Reward Function
- Implement the function inside the environment class using the reward signature.
- Use only existing attributes and methods from the provided environment code.
- Do not create or assume new fields beyond those defined.

3. Reward Design
- Shape rewards to encourage continuous progress toward success.
- Provide a large, one-time terminal reward when success is achieved.
- Use subtasks as guidance to design rewards.
- If the environment includes `self.task`, you may use it as an internal state 
  to avoid subtask conflicts.

4. Normalization and scaling
- Normalize each reward component into a bounded range within [0,1] before combining.
- Shaping functions (e.g. exp, tanh) may be used as part of producing this bounded form.
- After normalization, multiply each component by its explicit weight.
- The final reward is the sum of (weight × normalized_component) across all components.

5. Logging
- Define a dictionary named self.reward_info inside the function.
- Store each component with keys prefixed by "reward_" (e.g., "reward_distance").

## Output
Return only a single Python code block implementing the function:
```python
# your code here
```
\end{verbatim}
\end{tcolorbox}

\begin{tcolorbox}[breakable, colback=pastel_light_blue!10, colframe=pastel_light_blue!70!black, title=Reward Generation User Prompt]
\small
\begin{verbatim}
You are given the following input to design a reward function for a RL agent:
- task instruction: outlines the goal and intended agent behavior.
- Subtask list: defines sequential phases toward success.
- Environment code: includes the agent's observation and **success condition**.

Using this information, implement a reward function leading the agent to succeed task.

## Task Instruction
{task_instruction}

## Subtask List
{subtask_list}a

## Environment Code
{env_code}
\end{verbatim}
\end{tcolorbox}

\subsection{Trajectory Analysis}

\begin{tcolorbox}[breakable, colback=pastel_light_blue!10, colframe=pastel_light_blue!70!black, title=Trajectory Analysis System Prompt]
\small
\begin{verbatim}
You are a robotics evaluation analyst reviewing a trajectory produced by an RL policy. 
Your goal is to analyze the agent's behavior and evaluate the agent’s performance 
for each subtask using both visual observations and reward values.

## Instruction
1. Understand the task
Use the following inputs to infer the expected agent behavior and success criteria:
- Task Instruction: outlines the goal and intended agent behavior.
- Subtask list: defines sequential phases toward success.
- Reward function: reward function to train RL agent.
- Trajectory data per step, including:
  - Step index
  - Reward component values
  - Image (visual observation)

2. Evaluate subtask sequentially
Evaluate each subtask independently and in order.
- Use a 3-point success scale:
  - 1.0 → success
  - 0.5 → partial success
  - 0.0 → failure
- If a subtask fails, all later subtasks are assumed failed (0.0) unless there is 
a clear visual or reward-based evidence of later success.
- Support your evaluation with both image and reward evidence for each subtask.

3. Analysis Guidelines
For every subtask, perform two steps: Describe Behavior and Classify Success.

(i) Describe Behavior:
Observe key frames from the trajectory and describe how the agent behaves:
- What the agent is doing during this subtask.
- Check for unnatural or unintended behavior, such as:
  - Performing a valid subtask using an unintended strategy (e.g., crawling).
  - Exhibiting behaviors that negatively impact later subtasks (e.g., losing balance).

(ii) Classify Success
Evaluate the subtask outcome and assign a success level.
- Assign a score:
  - success (1.0), partial success (0.5), or failure (0.0).
- Provide concise reasoning for the assigned score:
  - Reference specific reward values at key steps (e.g., reward_lift = 0.92 at step 40).
  - Explain how these values justify the chosen success level.
  - If performance was poor or unstable, identify contributing factors, such as:
    - Reward gating or missing signal flow.
    - One component overshadowing others.
    - Reward magnitudes are too small or saturated.

## Output
For each subtask, report the following fields in structured JSON:
- `number`: subtask order
- `name`: subtask name
- `behavior`: description of how the agent behaves during this subtask.
- `score`: 3-point success score (0.0, 0.5, 1.0)
- `analysis`: concise reasoning that justifies the assigned score.

## Example
```json
{
  "subtasks": [
    {
      "number": 1,
      "name": "Stand up",
      "behavior": "The humanoid pushes off the ground with its hands, 
        extends its knees and reaches an upright posture with wobble. 
        No unintended behaviors observed.",
      "score": 1.0,
      "analysis": "Posture and stability rewards increased consistently 
      (reward_posture_upright = 0.95 at 15, reward_balance_stability = 0.90 at 18), 
      indicating full success in achieving upright balance."
    },
    {
      "number": 2,
      "name": "Move to package",
      "behavior": "The agent begins walking upright but transitions to a 
        crouched crawl midway, maintaining speed but reducing posture height.
        The unintended crawling compromises balance and future grasp alignment.",
      "score": 0.5,
      "analysis": "Forward velocity reward remained high 
        (reward_velocity_forward = 0.78 at step 35) 
        but posture reward dropped sharply 
        (reward_posture_upright = 0.38 at step 35),
        showing reliance on an unintended low-posture gait. 
        Partial success assigned due to progress despite unnatural locomotion."
    },
    ...
  ]
}
```
\end{verbatim}
\end{tcolorbox}

\begin{tcolorbox}[breakable, colback=pastel_light_blue!10, colframe=pastel_light_blue!70!black, title=Trajectory Analysis User Prompt]
\small
\begin{verbatim}
You are given the following information to evaluate a trajectory from a RL policy:
- Task Instruction: outlines the goal and intended agent behavior.
- Subtask list: defines sequential phases toward success.
- Reward function: reward function to train RL agent.
- Trajectory data per step, including:
  - Step index
  - Reward component values
  - Image (visual observation)

Your job is to evaluate each subtask in order. 
For every subtask, first describe the behavior from images.
Then classify success using reward values.

## Task Instruction
{task_instruction}

## Subtask List
{subtask_list}

## Reward Function
{reward_function}

## Trajectory
{trajectories}
\end{verbatim}
\end{tcolorbox}

\subsection{Subtask Reflection}

\begin{tcolorbox}[breakable, colback=pastel_light_blue!10, colframe=pastel_light_blue!70!black, title=Subtask Reflection System Prompt]
\small
\begin{verbatim}
You are a robotics task refinement analyst.
Your goal is to use the trajectory analysis to revise subtask definitions, 
which will later be used to design the reward function for RL.
The refinement should help agents exhibit more natural and goal-aligned behavior.

## Instruction

1. Understand the task
Use the following inputs to interpret both the intended and actual task execution:
- Task Instruction: outlines the goal and intended agent behavior.
- Subtask list: defines sequential phases toward success.
- Trajectory analysis: structured output containing, for each subtask:
  - behavior: description of what the agent did, including unnatural behavior.
  - score: success classification (1.0 = success, 0.5 = partial, 0.0 = failure).
  - analysis: reasoning based on evidence and causal factors behind success or failure.

2. Refine the subtasks
Based on your understanding of the trajectory analysis:
- Determine whether refinement is needed.
- If no refinement is necessary, keep the original subtask list.
- If refinement is needed, select one subtask that most contributed to unintended behavior.
  - Revise the subtask to encourage natural and intended behavior.
  - Avoid numeric thresholds.
  - Keep the subtasks concise.
  - Keep the same total number of subtasks.

## Output

Return a JSON object containing:
- `decision`: A short paragraph explaining which subtask (if any) was refined, 
    why or why not, and how the change supports better reward function design.
- `subtasks`: The final subtask list, unchanged if no refinement is needed,
    or with one updated definition if refinement was applied.

### Example

```json
{
  "decision": "Subtask 2 ('Move to package') was refined 
    because the agent crawled instead of walking.
    The refined version enforces upright walking 
    to align the behavior with intended locomotion.",
  "subtasks": [
    "1. Stand upright with stable balance.",
    "2. Walk upright toward the package.",
    "3. Grasp and lift the package smoothly."
    "4. Place the package on the table at the target location as close as possible."
  ]
}
```
\end{verbatim}
\end{tcolorbox}

\begin{tcolorbox}[breakable, colback=pastel_light_blue!10, colframe=pastel_light_blue!70!black, title=Subtask Reflection User Prompt]
\small
\begin{verbatim}
You are given the following to refine a subtask list for an RL task:
- Task Instruction: outlines the goal and intended agent behavior.
- Subtask list: defines sequential phases toward success.
- Trajectory analysis: per-subtask results from the evaluation stage
(includes behavior, score, and reward-based analysis)

Your job is to decide whether any refinement is needed.
If needed, revise one subtask to encourage natural and intended behavior.
If no refinement is needed, keep the list unchanged.

## Task Instruction
{task_instruction}

## Subtask List
{subtask_list}

## Trajectory Analysis
{traj_analysis_summary}
\end{verbatim}
\end{tcolorbox}

\subsection{Reward Reflection}

\begin{tcolorbox}[breakable, colback=pastel_light_blue!10, colframe=pastel_light_blue!70!black, title=Reward Reflection System Prompt]
\small
\begin{verbatim}
You are a reward engineering expert assisting reinforcement learning agents.
Your goal is to revise an existing reward function to maximize task success.

## Instructions

1. Understand the context
Use the following inputs to understand the context of the task:
- Task description: outlines the goal and intended agent behavior.
- Subtask list: defines sequential phases toward success.
- Environment code: includes the agent's observation and **success condition**.
- Previous reward function and its analysis:
  - reward function code.
  - task success rate (from the model’s reflection).
  - trajectory analysis: evaluation of agent behavior.

2. Revise the Reward Function
- Revise the reward function inside the environment class using the exact signature:
```
{reward_signature}.
```
- Use only existing attributes and methods from the provided environment code.
- Do not create or assume new fields beyond those defined.

3. Revision Strategy
- Make key changes that directly address the reported failure(s).
- If all subtasks are failing, you may rewrite the reward function entirely.
- You may re-weight existing components as part of the revision.

4. Reward Design
- Shape rewards to encourage continuous progress toward success.
- Provide a large, one-time terminal reward when success is achieved 
    (success function returns True).
- Use subtasks as guidance to design rewards.
- If the environment includes `self.task`, 
    you may use it as an internal state to avoid subtask conflicts.

5. Normalization and scaling
- Normalize each reward component into a bounded range within [0,1] before combining.
- Shaping functions (e.g. exp, tanh) may be used as part of producing this bounded form.
- After normalization, multiply each component by its explicit weight.
- The final reward is the sum of (weight × normalized_component) across all components.

6. Logging
- Define a dictionary named self.reward_info inside the function.
- Store each component with keys prefixed by "reward_" (e.g., "reward_distance").

## Output

1. Revision notes
Briefly summarize what was changed and which failures or low-performing subtasks.

2. Revised Reward Function
Return a single Python code block implementing the function:
```python
# your code here
```
\end{verbatim}
\end{tcolorbox}

\begin{tcolorbox}[breakable, colback=pastel_light_blue!10, colframe=pastel_light_blue!70!black, title=Reward Reflection User Prompt]
\small
\begin{verbatim}
You are given the following information to revise a reward function for a RL agent:
- Task description: outlines the goal and intended agent behavior.
- Subtask list: defines sequential phases toward success.
- Environment code: includes agent's observation and **success condition**.
- Previous reward function and its analysis, including:
  - reward function code.
  - task success rate (from the model’s reflection).
  - trajectory analysis describing behavior.

Your job is to revise the reward function by introducing one modification 
that addresses key failures.
The revised reward should improve the likelihood of task success.

## Task Instruction
{task_instruction}

## Subtask List
{subtask_list}

## Environment Code
{env_code}

## Previous Reward Function and its Analysis

### Previous Reward Function
{reward_function}

### Success Rate
{self_task_success}

### Trajectory Analysis
{traj_analysis_summary}
\end{verbatim}
\end{tcolorbox}

\newpage

\section{RDA Examples}
\label{appendix:rda_example}

To illustrate the RDA procedure concretely, we present a detailed example for 3 tasks showing one complete iteration cycle. Specifically, we provide the following components:

\textbf{I. Initialization:}
\begin{itemize}
    \item Task instruction $I$
    \item Environment Code $E$
    \item Initial subtask decomposition $T^1 = [t_1^1, t_2^1, \ldots, t_J^1]$
    \item Initial reward function for candidate $n$: $R_n^1$
\end{itemize}
\textbf{II. Evolutionary Search:}
\begin{itemize}
    \item Example trajectory from the trained policy rollout: $\tau_{n,k}^1$.
    \item Trajectory analysis report: $A_{n,k}^1.$
    \item Policy summary: $G_n^1.$
    \item Refined subtask decomposition $T^2,$
    \item Refined reward function for candidate $n$: $R_n^2.$
\end{itemize}

\subsection{(ManiSkill) PlugCharger}

\begin{tcolorbox}[breakable, colback=pastel_light_blue!10, colframe=pastel_light_blue!70!black, title=Task Instruction]
\small
\begin{verbatim}
### Instruction
This env requires the robot to pick up the charger and insert it into the receptacle.

The env is considered successful when the following conditions are met:
- The charger is inserted into the receptacle.
\end{verbatim}
\end{tcolorbox}

\begin{tcolorbox}[breakable, colback=pastel_light_blue!10, colframe=pastel_light_blue!70!black, title=Environment Code]
\small
\begin{verbatim}
Observations contain both the robot and the objects it manipulates.

The environment includes randomization to ensure robust learning:    
- The charger position is randomized on the XY plane on top of the table.
The rotation is also randomized 
- The receptacle position is randomized on the XY plane and the rotation is randomized.
Note that the human render camera has its pose fixed relative to the receptacle.

@register_env("PlugCharger-v1", max_episode_steps=200)
class PlugChargerEnv:
    """
    Functions:
    - `_load_agent(self)`: Initializes the robot and sets its physical properties.
    - `_load_scene(self)`: Loads the scene, including objects and environment layout.
    - `_get_obs_extra(self)`: Defines the observation space returned to the agent.
    - Access extra obs information via obs['extra'][attribute_name].
    - `evaluate(self)`: Assesses whether the environment has reached a success condition.

    Attributes:
    - `agent`: Instance of the robot, represented by the `Agent` class.
    - `agent.tcp.pose.p`: End-effector position of the robot.
    - `agent.tcp.pose.q`: End-effector orientation (quaternion).
    - `agent.is_grasping(obj)`: Indicates whether the robot is holding an object.
       Note: is_grasping(obj) is not available when using the Panda stick agent.

    - `obj`: Environment object being manipulated.
    - `obj.pose.p`: Position of the object, from the `Pose` class.
    - `obj.pose.q`: Orientation of the object (quaternion), from the `Pose` class.
    - `obj.pose.inv()`: Inverse pose of the object, useful for relative poses.

    Any object that is an instance of the `Pose` class has `.p`, `.q`, and `.inv()`
    [Warning]                                                    
    - `agent.is_grasping(obj)`, `.p`, `.q`, and `.inv()` are `torch.Tensor`.
    - Specifically, `.p` has shape (n, 3) and `.q` has shape (n, 4).
    - To access scalar values, use indexing (e.g., `tensor[i, j]`) or `.item()`.
    - Do not use `int(tensor)`, `tensor.int()`, `float(tensor)`, `tensor.float()`.
    """
    _base_size = [2e-2, 1.5e-2, 1.2e-2]
    _peg_size = [8e-3, 0.75e-3, 3.2e-3]
    _peg_gap = 7e-3
    _clearance = 5e-4
    _receptacle_size = [1e-2, 5e-2, 5e-2]
    SUPPORTED_ROBOTS = ["panda_wristcam"]
    agent: Union[PandaWristCam]

    def __init__(
        self, *args, robot_uids, robot_init_qpos_noise, **kwargs
    ):
        self.robot_init_qpos_noise = robot_init_qpos_noise
        super().__init__(*args, robot_uids=robot_uids, **kwargs)

    def _get_obs_extra(self, info: Dict):
        obs = dict(tcp_pose=self.agent.tcp.pose.raw_pose)
        if self.obs_mode_struct.use_state:
            obs.update(
                charger_pose=self.charger.pose.raw_pose,
                receptacle_pose=self.receptacle.pose.raw_pose,
                goal_pose=self.goal_pose.raw_pose,
            )
        return obs

    def evaluate(self):
        obj_to_goal_dist, obj_to_goal_angle = self._compute_distance()
        success = (obj_to_goal_dist <= 5e-3) & (obj_to_goal_angle <= 0.2)
        return dict(
            obj_to_goal_dist=obj_to_goal_dist,
            obj_to_goal_angle=obj_to_goal_angle,
            success=success,
        )
\end{verbatim}
\end{tcolorbox}

\begin{tcolorbox}[breakable, colback=pastel_light_blue!10, colframe=pastel_light_blue!70!black, title=Initial Subtask]
\small
\begin{verbatim}
1. Move the end-effector above the charger at a grasp-ready pose.
2. Lower and close the gripper to grasp the charger.
3. Lift the charger off the table.
4. Move to the receptacle and align the charger’s plug with the receptacle opening.
5. Insert the charger into the receptacle.
\end{verbatim}
\end{tcolorbox}

\begin{tcolorbox}[breakable, colback=pastel_light_blue!10, colframe=pastel_light_blue!70!black, title=Initial Reward Funcion]
\small
\begin{verbatim}
```python
def compute_dense_reward(self, obs: Any, action: torch.Tensor) -> torch.Tensor:
    # Gather poses
    tcp_p = self.agent.tcp.pose.p  # (n, 3)
    tcp_q = self.agent.tcp.pose.q  # (n, 4)
    charger_p = self.charger.pose.p  # (n, 3)
    charger_q = self.charger.pose.q  # (n, 4)

    # Useful scalars and masks
    n = tcp_p.shape[0]
    eval_dict = self.evaluate()
    obj_to_goal_dist = eval_dict["obj_to_goal_dist"]  # (n,)
    obj_to_goal_angle = eval_dict["obj_to_goal_angle"]  # (n,)
    success_mask = eval_dict["success"]  # (n,) bool
    grasp_mask = self.agent.is_grasping(self.charger)  # (n,) bool
    grasp_f = grasp_mask.to(dtype=torch.float32)
    success_f = success_mask.to(dtype=torch.float32)

    # Helper: quaternion angular difference via inner product (robust to sign)
    # angle in [0, pi]; q are unit quaternions
    q_dot_tcp_charger = torch.sum(tcp_q * charger_q, dim=-1).abs().clamp(0.0, 1.0)
    ang_tcp_charger = 2.0 * torch.acos(q_dot_tcp_charger)

    # Distances
    d_tcp_charger = torch.norm(tcp_p - charger_p, dim=-1)  # (n,)
    dz_tcp_charger = torch.abs(tcp_p[:, 2] - charger_p[:, 2])  # (n,)

    # Normalization scales (chosen based on typical tabletop manipulation ranges)
    s_approach_pos = 0.25  # meters
    s_approach_z = 0.05    # meters
    s_approach_ang = 0.5   # radians
    s_goal_dist = 0.05     # meters
    s_goal_ang = 0.4       # radians
    s_hold_close = 0.01    # meters

    # Normalized components in [0, 1]
    r_approach_pos = torch.exp(-d_tcp_charger / s_approach_pos)  # TCP to charger
    r_approach_z = torch.exp(-dz_tcp_charger / s_approach_z) # closer in vertical
    r_approach_ang = torch.exp(-ang_tcp_charger / s_approach_ang)  # align TCP

    r_grasp = grasp_f  # binary: 1 when grasped

    # Once grasped, encourage moving the charger toward receptacle goal 
    # and aligning the orientation
    r_goal_dist = torch.exp(-obj_to_goal_dist / s_goal_dist) * grasp_f
    r_goal_ang = torch.exp(-obj_to_goal_angle / s_goal_ang) * grasp_f

    # Encourage keeping the charger close to TCP when grasped (stability)
    r_hold_close = torch.exp(-d_tcp_charger / s_hold_close) * grasp_f

    # Large terminal reward on success
    r_success = success_f  # binary, normalized

    # Weights for each component
    w_approach_pos = 1.0
    w_approach_z = 0.3
    w_approach_ang = 0.5
    w_grasp = 1.5
    w_goal_dist = 2.0
    w_goal_ang = 1.0
    w_hold_close = 0.5
    w_success = 8.0

    # Combine components
    reward = (
        w_approach_pos * r_approach_pos
        + w_approach_z * r_approach_z
        + w_approach_ang * r_approach_ang
        + w_grasp * r_grasp
        + w_goal_dist * r_goal_dist
        + w_goal_ang * r_goal_ang
        + w_hold_close * r_hold_close
        + w_success * r_success
    )

    # Logging
    self.reward_info = {
        "reward_approach_charger_pos": w_approach_pos * r_approach_pos,
        "reward_approach_charger_height": w_approach_z * r_approach_z,
        "reward_approach_charger_orient": w_approach_ang * r_approach_ang,
        "reward_grasp": w_grasp * r_grasp,
        "reward_move_to_goal_dist": w_goal_dist * r_goal_dist,
        "reward_align_to_goal_orient": w_goal_ang * r_goal_ang,
        "reward_hold_close": w_hold_close * r_hold_close,
        "reward_success": w_success * r_success,
        "reward_total": reward,
    }

    return reward
\end{verbatim}
\end{tcolorbox}

\begin{figure}[h]
\begin{center}
\includegraphics[width=.99\textwidth]{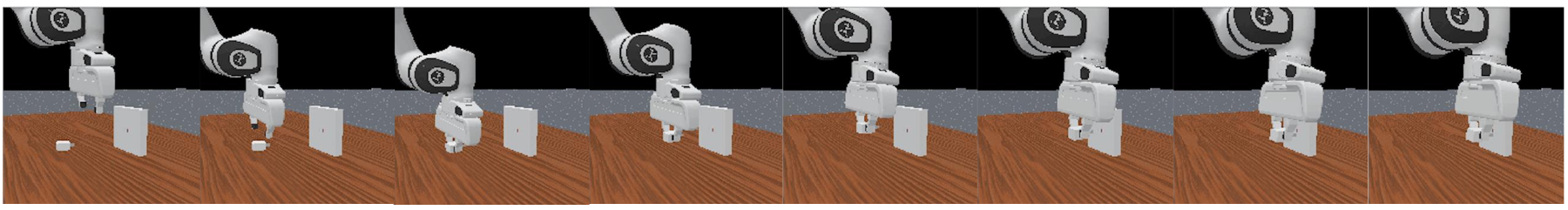}
\end{center}
\vspace{-2mm}
\caption{\textbf{PlugCharger Trajectory.}
Illustration of a PlugCharger trajectory generated by the policy, trained with the initial example reward function.}
\label{figure:plug_charger_supp}
\end{figure}

\begin{tcolorbox}[breakable, colback=pastel_light_blue!10, colframe=pastel_light_blue!70!black, 
title=Trajectory Analysis]
\small
\begin{verbatim}
{
  "1": {
    "number": 1,
    "name": "Move the end-effector above the charger at a grasp-ready pose.",
    "behavior": "From the initial view, the gripper approaches the charger. 
        By the early steps, the end-effector is positioned close to the charger 
        and appears ready to descend for a top-down grasp.",
    ,
    "score": 1.0,
    "analysis": "Early approach and height signals increase appropriately: 
        reward_approach rises from 0.098 (step 2) to 0.366 (step 6), and reward_above 
        is high at 0.199 (step 2) before decreasing as the gripper prepares to descend. 
        These values indicate EE moved near and above the charger in grasp-ready.",
  },
  "2": {
    "number": 2,
    "name": "Lower and close the gripper to grasp the charger.",
    "behavior": "The gripper lowers toward the charger and closes. 
        After contact, the agent maintains a hold 
        while beginning to move away from the table surface.",
    "score": 1.0,
    "analysis": "Grasp is clearly achieved: 
        reward_grasp switches to 1.000 at step 8 and remains at 1.000 afterward. 
        The decrease in reward_above (e.g., 0.075 at step 8) is consistent 
        with lowering to make contact. 
        This provides strong evidence of a successful grasp.",
  },
  "3": {
    "number": 3,
    "name": "Lift the charger off the table.",
    "behavior": "With the gripper closed, the agent lifts the charger, 
        moving it to approximately the receptacle height. 
        The grasp remains stable throughout.",
    "score": 1.0,
    "analysis": "Lift reward increases immediately after grasp: 
        reward_lift goes from 0.012 (step 8) to 0.329 (step 10) and stabilizes near ~0.47 
        (e.g., 0.477 at step 12, 0.470 at step 40),
        showing the object is raised toward the goal height while being held.",
  },
  "4": {
    "number": 4,
    "name": "Move to the receptacle and align the charger’s plug with the receptacle opening.",
    "behavior": "The agent carries the charger to the receptacle face and holds position. 
        Visually, the gripper stays near the receptacle across many frames 
        (e.g., steps 10, 20, 40, 52), with little motion into the opening.",
    "score": 0.5,
    "analysis": "Positional proximity becomes and stays moderate: 
        reward_goal_pos_post increases to 0.364 at step 10 and plateaus at 0.714 
        from step 14 onward, indicating the charger is close 
        but not centered at the opening (distance not 0). 
        Orientation is very good: 
        reward_goal_ang_post ~0.48 (e.g., 0.480\u20130.484 at steps 40\u2013200), 
        implying near-perfect angular alignment gated by positional proximity. 
        Partial success is assigned because alignment is strong, 
        but the lateral centering/approach remains offset by a centimeter, 
        so the plug is near but not precisely at the opening.",
  },
  "5": {
    "number": 5,
    "name": "Insert the charger into the receptacle.",
    "behavior": "The agent maintains a hover in front of the receptacle 
        without pushing the charger. 
        Across the later frames, there is no visible insertion. 
        The gripper remains stationary, and the episode ends without success.",
    "score": 0.0,
    "analysis": "No terminal success occurs: 
        reward_success_bonus remains 0.000 throughout. 
        Despite sustained proximity and alignment (reward_goal_pos_post ~0.722\u20130.724), 
        the agent does not cross the final insertion threshold. 
        Likely factors: the dense shaping saturates near the receptacle 
        (goal_pos_norm_post ~0.72), providing limited incentive to push further; 
        insertion relies on a sparse success bonus (10.0) without contact shaping, 
        making the final step to encourage the policy to hover at a high-reward plateau.",
  }
}
\end{verbatim}
\end{tcolorbox}

\begin{tcolorbox}[breakable, colback=pastel_light_blue!10, colframe=pastel_light_blue!70!black, title=Subtask Reflection]
\small
\begin{verbatim}
"decision": "Subtask 4 was refined because the agent stalled 
    while hovering near the opening 
    with saturated proximity/orientation rewards, 
    never transitioning to pre-insertion. 
    
    The original wording allowed "ready to insert" to be satisfied by aligned poses,
    which encouraged hovering. 
    
    The refined subtask requires gentle contact at the opening with a centered tip 
    and axial alignment, enabling reward shaping for contact-based readiness 
    and discouraging indecisive hovering.",

"subtasks": [
    "1. Move the end-effector above the charger at a grasp-ready pose.",
    "2. Lower and close the gripper to grasp the charger.",
    "3. Lift the charger off the table.",
    "4. Bring the charger to the receptacle opening; center the plug tip, 
        align the insertion axis, and make gentle, stable contact 
        at the opening to establish a pre-insertion pose.",
    "5. Insert the charger into the receptacle."
]
\end{verbatim}
\end{tcolorbox}

\begin{tcolorbox}[breakable, colback=pastel_light_blue!10, colframe=pastel_light_blue!70!black, title=Reward Reflection]
\small
\begin{verbatim}
Revision notes
- The agent was aligning orientation “in the air.” 
without translating the charger toward the receptacle,
due to an overly steep positional shaping term (exp(-200·dist)) 
that yielded near-zero rewards when more than a few centimeters away.
- Two tightly related modifications:
  1) Soften the positional shaping for charger-to-receptacle distance 
    by reducing the exponent coefficient from 200 to 20 
    to maintain a usable gradient over larger distances.
  2) Gate the post-grasp orientation reward by positional proximity 
    (multiply orientation alignment by the post-grasp position proximity term) 
    so orientation is rewarded primarily when the charger is actually 
    near the receptacle, discouraging alignment far from the opening.

These changes directly target the failure in subtask 4 (lack of translation), 
which in turn improves the likelihood of completing subtask 5 (insertion).

```python
def compute_dense_reward(self, obs: Any, action: torch.Tensor) -> torch.Tensor:

    ...
    
    # Near-goal insertion-drive component (sharply increases within final centimeter),
    # gated by grasp and alignment to encourage pushing straight into the slot.
    precision_scale = 0.01  # meters (focus gradient on final push)
    reward_insert_drive_norm = (
        torch.exp(-obj_to_goal_dist / precision_scale).clamp(0.0, 1.0)
        * reward_align_norm
        * gate
    )  # stays within [0,1]

    # Weights for each component
    w_hold_close = 0.3 # slightly reduced to avoid overshadowing final push
    w_insert_drive = 2.0  # emphasize precision near-goal progress

    ...
    
    return reward
\end{verbatim}
\end{tcolorbox}

\newpage

\subsection{(HumanoidBench) Package Delievery}

\begin{tcolorbox}[breakable, colback=pastel_light_blue!10, colframe=pastel_light_blue!70!black, title=Task Instruction]
\small
\begin{verbatim}
The humanoid begins each episode standing upright, maintaining balance.
From this stable stance, it must walk steadily toward the box.
Upon reaching the box, the humanoid positions its both hands behind it, 
establishing a firm contact with the surface.
The box must then be pushed while the humanoid remains balanced and maintains momentum.
Throughout the push, the humanoid regulates the applied force to move the box smoothly 
toward the designated target location without losing posture or control.
The task succeeds when the box is accurately positioned at the target location.
\end{verbatim}
\end{tcolorbox}

\begin{tcolorbox}[breakable, colback=pastel_light_blue!10, colframe=pastel_light_blue!70!black, title=Environment Code]
\small
\begin{verbatim}
from typing import Dict, Tuple
import numpy as np
from gymnasium.spaces import Box
from humanoid_bench.tasks import Task

ROBOT_STAND_HEIGHT = 1.65
PACKAGE_HALF_SIZE = np.array([0.4, 0.4, 0.4])
PACKAGE_POSITION = np.array([2.0, -0.5, 0.4])
PACKAGE_SIZE = PACKAGE_HALF_SIZE * 2
PACKAGE_DENSITY = 5.0
PACKAGE_VOLUME = np.prod(PACKAGE_SIZE)
PACKAGE_MASS = PACKAGE_VOLUME * PACKAGE_DENSITY


class H1Hand:
    dof = 76

    # Global Variables
    def joint_angles(self): ...  # shape (dof-7,)
    def joint_velocities(self): ...  # shape (dof-7,)
    def control(self): ...  # shape (n_ctrl,)
    def actuator_forces(self): ...  # shape (n_actuators,)

    # Head
    def head_position(self): ...  # shape (3,)
    def head_height(self): ...  # scalar ()

    # Shoulders
    def left_shoulder_position(self): ...  # shape (3,)
    def left_shoulder_orientation(self): ...  # shape (9,)
    def right_shoulder_position(self): ...  # shape (3,)
    def right_shoulder_orientation(self): ...  # shape (9,)

    # Elbows
    def left_elbow_position(self): ...  # shape (3,)
    def left_elbow_orientation(self): ...  # shape (9,)
    def right_elbow_position(self): ...  # shape (3,)
    def right_elbow_orientation(self): ...  # shape (9,)

    # Hands
    def left_hand_position(self): ...  # shape (3,)
    def left_hand_orientation(self): ...  # shape (9,)
    def left_hand_velocity(self): ...  # shape (3,)
    def right_hand_position(self): ...  # shape (3,)
    def right_hand_orientation(self): ...  # shape (9,)
    def right_hand_velocity(self): ...  # shape (3,)

    # Torso
    def torso_position(self): ...  # shape (3,)
    def torso_orientation(self): ...  # shape (9,)
    def torso_upright(self): ...  # scalar ()
    def torso_vertical_orientation(self): ...  # shape (3,)

    # Pelvis
    def pelvis_position(self): ...  # shape (3,)
    def pelvis_orientation(self): ...  # shape (9,)
    def center_of_mass_position(self): ...  # shape (3,)
    def center_of_mass_velocity(self): ...  # shape (3,)
    def body_velocity(self): ...  # shape (3,)

    # Knees
    def left_knee_position(self): ...  # shape (3,)
    def left_knee_orientation(self): ...  # shape (9,)
    def right_knee_position(self): ...  # shape (3,)
    def right_knee_orientation(self): ...  # shape (9,)

    # Feet
    def left_foot_position(self): ...  # shape (3,)
    def left_foot_height(self): ...  # scalar ()
    def left_foot_orientation(self): ...  # shape (9,)
    def right_foot_position(self): ...  # shape (3,)
    def right_foot_height(self): ...  # scalar ()
    def right_foot_orientation(self): ...  # shape (9,)
    

class Package(Task):
    def __init__(self, robot=None, env=None, **kwargs):
        super().__init__(robot, env, **kwargs)

    def observation_space(self):
        return Box(
            low=-np.inf,
            high=np.inf,
            shape=(self.robot.dof * 2 - 1 + self.dof * 2 - 1 + 9,),
            dtype=np.float64,
        )

    def goal_dist(self, ...):
        pass

    def get_obs(self):
        position = self._env.data.qpos.flat.copy()[: self.robot.dof]
        velocity = self._env.data.qvel.flat.copy()[: self.robot.dof - 1]
        package_destination = self._env.named.data.site_xpos["destination_loc"]
        box_pos = self._env.data.qpos.flat.copy()[-7:]
        box_vel = self._env.data.qvel.flat.copy()[-6:]
        left_hand = self.robot.left_hand_position()
        right_hand = self.robot.right_hand_position()

        return np.concatenate(
            [
                position,
                velocity,
                package_destination,
                box_pos,
                box_vel,
                left_hand,
                right_hand,
            ]
        )

    def get_success(self):
        package_destination = self._env.named.data.site_xpos["destination_loc"]
        package_location = self._env.named.data.qpos["free_package"][:3]

        dist_package_destination = np.linalg.norm(
            package_location[:2] - package_destination[:2]
        )
        success = dist_package_destination < 0.1
        return success, {
            "success": success,
        }

    def get_terminated(self):
        success = self.get_success()[0]

        # Check for collapse
        pelvis_height = self.robot.pelvis_position()[2]  # z-coordinate of pelvis
        fail = pelvis_height < 0.5

        terminated = success or fail
        return terminated, {}

    def reset_model(self):
        q_pos = self._env.data.qpos.copy()
        q_pos[-7] = PACKAGE_POSITION[0]  # x
        q_pos[-6] = PACKAGE_POSITION[1]  # y
        q_pos[-5] = PACKAGE_POSITION[2]  # z

        q_vel = self._env.data.qvel.copy()
        self._env.set_state(q_pos, q_vel)

        # --- RANDOMIZE GOAL BODY POSITION ---
        base_goal_position = np.array([2.0, 1.5, 0.4])
        goal_position = base_goal_position + np.array(
            [np.random.uniform(-0.5, 0.5), np.random.uniform(-0.5, 0.5), 0.0]
        )
        self._env.model.body_pos[-1] = goal_position
        return super().reset_model()
\end{verbatim}
\end{tcolorbox}

\begin{tcolorbox}[breakable, colback=pastel_light_blue!10, colframe=pastel_light_blue!70!black, title=Initial Subtask]
\small
\begin{verbatim}
1. Stabilize in an upright stance.
2. Walk to the box and position behind it.
3. Place both hands on the rear face of the box, establishing firm contact.
4. Align the body and box to face the destination.
5. Push the box forward while maintaining balance and steady momentum.
6. Continue pushing, modulating force to move the box smoothly toward the destination.
7. Fine-position and stop pushing when the box is within 0.1 m of the destination location.
\end{verbatim}
\end{tcolorbox}

\begin{tcolorbox}[breakable, colback=pastel_light_blue!10, colframe=pastel_light_blue!70!black, title=Initial Reward Funcion]
\small
\begin{verbatim}
```python
def get_reward(self) -> Tuple[float, Dict]:
    # Helper functions
    def clip01(x):
        return float(np.clip(x, 0.0, 1.0))

    def exp_norm(x, scale):
        # Returns exp(-|x|/scale) in [0,1], larger when x is small
        return float(np.exp(-abs(x) / max(scale, 1e-6)))

    def exp_norm_sq(x, scale):
        # Returns exp(-(x/scale)^2) in [0,1], larger when x is small
        return float(np.exp(- (x / max(scale, 1e-6)) ** 2))

    # Read current state
    success, _ = self.get_success()
    destination = self._env.named.data.site_xpos["destination_loc"]
    box_qpos = self._env.data.qpos.flat.copy()[-7:]
    box_pos = box_qpos[:3]
    box_vel = self._env.data.qvel.flat.copy()[-6:]
    box_lin_vel = box_vel[:3]
    box_ang_vel = box_vel[3:]

    pelvis_pos = self.robot.pelvis_position()
    torso_pos = self.robot.torso_position()
    torso_upright_val = self.robot.torso_upright()

    left_hand = self.robot.left_hand_position()
    right_hand = self.robot.right_hand_position()

    # Geometry toward destination
    dir_vec = destination - box_pos
    dist_box_to_dest = float(np.linalg.norm(dir_vec))
    dir_hat = dir_vec / (dist_box_to_dest + 1e-8)

    # Subtask 1: Stabilize upright stance
    # Height close to stand height
    pelvis_height = float(pelvis_pos[2])
    height_err = abs(pelvis_height - ROBOT_STAND_HEIGHT)
    reward_upright_height_norm = exp_norm_sq(height_err, scale=0.20) 
    # Torso upright indicator (assumed in [0,1], clipped)
    reward_upright_torso_norm = clip01(torso_upright_val)

    # Subtask 2: Walk to the box and position behind it (relative to destination)
    behind_offset = PACKAGE_HALF_SIZE[0] + 0.50  # half-size + clearance
    target_behind_pos = box_pos - dir_hat * behind_offset
    dist_to_behind = float(np.linalg.norm(torso_pos - target_behind_pos))
    reward_approach_behind_norm = float(np.exp(-dist_to_behind / 2.0))
    # Ensure torso is indeed behind the box along destination axis
    behind_sign = float(np.dot(dir_hat, box_pos - torso_pos)) 
    reward_behind_alignment_norm = 0.5 * (1.0 + np.tanh(behind_sign / 0.30))

    # Subtask 3: Place both hands on the rear face of the box
    def hand_contact_norm(hand_pos):
        rel = hand_pos - box_pos
        parallel = float(np.dot(rel, dir_hat))  # along destination axis; 
        perp_vec = rel - parallel * dir_hat
        radial = float(np.linalg.norm(perp_vec)) # distance from axis center
        vertical_diff = abs(float(hand_pos[2] - box_pos[2]))
        plane_norm = exp_norm(abs(parallel + PACKAGE_HALF_SIZE[0]), scale=0.10)  
        radial_norm = exp_norm(radial, scale=0.30) # centered laterally
        vertical_norm = exp_norm(vertical_diff, scale=0.20) # near center height
        # Combine multiplicatively to enforce simultaneous proximity
        return float(np.clip(plane_norm * radial_norm * vertical_norm, 0.0, 1.0))

    left_contact = hand_contact_norm(left_hand)
    right_contact = hand_contact_norm(right_hand)
    reward_hand_contact_norm = float((left_contact + right_contact) / 2.0)

    # Subtask 4: Align body and box to face destination (torso behind)
    # Use vector torso->box alignment with dest direction
    v_tb = box_pos - torso_pos
    v_tb_norm = float(np.linalg.norm(v_tb))
    v_tb_hat = v_tb / (v_tb_norm + 1e-8)
    alignment_val = float(np.dot(v_tb_hat, dir_hat))  # in [-1,1]
    reward_body_dest_align_norm = clip01(alignment_val)  # favor [0,1]

    # Subtask 5, 6: Push the box forward while maintaining balance 
    # and modulate force for smooth motion
    forward_speed = float(np.dot(box_lin_vel, dir_hat))  # m/s along destination
    reward_forward_speed_norm = clip01(forward_speed / 0.60)  # normalize to target
    # Prefer a moderate target speed ~0.4 m/s
    reward_speed_match_norm = exp_norm_sq(forward_speed - 0.40, scale=0.40)
    # Discourage rotation while pushing (smooth motion)
    ang_speed = float(np.linalg.norm(box_ang_vel))
    reward_angular_stability_norm = exp_norm(ang_speed, scale=1.0)

    # Subtask 6 continued: Progress toward destination via distance shaping
    reward_distance_progress_norm = 1.0 - float(np.tanh(dist_box_to_dest / 3.0))  

    # Subtask 7: Fine positioning near goal and stop pushing
    near_goal_factor = float(np.exp(-max(dist_box_to_dest - 0.10, 0.0) / 0.20))
    reward_near_goal_stop_norm = near_goal_factor * exp_norm_sq(forward_speed, 0.2)

    # Terminal success reward
    reward_success_norm = 1.0 if success else 0.0

    # only reward pushing strongly when hands are in contact and positioned behind
    contact_gate = clip01(reward_hand_contact_norm * reward_behind_alignment_norm)

    # Weights for each component
    w_upright_height = 0.15
    w_upright_torso = 0.15
    w_approach_behind = 0.15
    w_behind_alignment = 0.05
    w_hand_contact = 0.15
    w_body_dest_align = 0.05
    w_distance_progress = 0.15
    w_forward_speed = 0.10
    w_speed_match = 0.05
    w_angular_stability = 0.05
    w_near_goal_stop = 0.05
    w_success = 8.0  # large terminal bonus

    # Combine rewards (weights * normalized components)
    reward_components = {
"reward_upright_height": w_upright_height * reward_upright_height_norm,
"reward_upright_torso": w_upright_torso * reward_upright_torso_norm,
"reward_approach_behind": w_approach_behind * reward_approach_behind_norm,
"reward_behind_alignment": w_behind_alignment * reward_behind_alignment_norm,
"reward_hand_contact": w_hand_contact * reward_hand_contact_norm,
"reward_body_dest_align": w_body_dest_align * reward_body_dest_align_norm,
"reward_distance_progress": w_distance_progress * reward_distance_progress_norm,
"reward_forward_speed": w_forward_speed * (reward_forward_speed_norm * contact_gate),
"reward_speed_match": w_speed_match * (reward_speed_match_norm * contact_gate),
"reward_angular_stability": w_angular_stability * reward_angular_stability_norm,
"reward_near_goal_stop": w_near_goal_stop * reward_near_goal_stop_norm,
"reward_success": w_success * reward_success_norm,
    }

    reward = float(sum(reward_components.values()))

    # Logging
    self.reward_info = dict(reward_components)
    self.reward_info.update({
        "reward_total": reward,
        "success": success,
        "dist_box_to_destination": dist_box_to_dest,
        "forward_speed_along_dest": forward_speed,
    })

    reward_info = self.reward_info
    return reward, reward_info
```
\end{verbatim}
\end{tcolorbox}

\begin{figure}[h]
\begin{center}
\includegraphics[width=.99\textwidth]{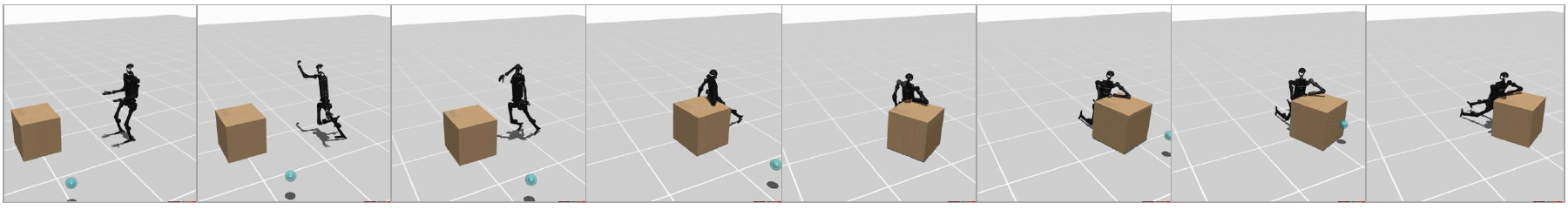}
\end{center}
\vspace{-2mm}
\caption{\textbf{Package Delivery Trajectory.}
Illustration of a Package Delivery trajectory generated by the policy, trained with the initial example reward function.}
\label{figure:humanoid_package_supp}
\end{figure}

\begin{tcolorbox}[breakable, colback=pastel_light_blue!10, colframe=pastel_light_blue!70!black, 
title=Trajectory Analysis]
\small
\begin{verbatim}
{
  "1": {
    "number": 1,
    "name": "Stabilize in an upright stance",
    "behavior": "The humanoid begins roughly upright but quickly collapses 
        into a crouched posture. 
        By step 25, the body is pitched forward toward the box, 
        and throughout the rollout, it maintains a low, 
        hunched stance rather than an upright balance."
    "score": 0.5,
    "analysis": "Upright reward starts moderate then drops, 
        indicating loss of upright stability: 
        reward_upright = 0.413 at step 0, 0.265 at step 10, 
        0.105 at step 35, and 0.091 at step 50. 
        This shows only brief initial uprightness followed by a sustained crouch. 
        Partial credit for the initial momentary upright stance, 
        but it is not maintained.",
  },
  "2": {
    "number": 2,
    "name": "Walk to the box and position behind it",
    "behavior":  "The agent approaches the box and moves to its rear side 
        relative to the destination. 
        Visuals show the robot coming up to the back of the box.",
    "score": 1.0,
    "analysis":
      "Approach improves consistently: 
        reward_approach_box = 0.443 at step 20, 0.573 at step 35, 0.584 at step 40.
        Behindness rises from 0.112 at step 0 to 0.578 at step 35 and 0.952 at step 50, 
        confirming it has positioned behind the box as intended.",
  },
  "3": {
    "number": 3,
    "name": "Place both hands on the rear face of the box, establishing firm contact",
    "behavior":  "The robot gets its upper body against the rear face and makes 
        intermittent or uneven hand contact while leaning on the box. 
        Contact is present but not consistently strong or symmetric.",
    "score": 0.5,
    "analysis":  "Hand-contact signal is present but moderate and variable:
        reward_hand_contact = 0.143 at step 35, 0.247 at step 40, 0.318 at step 45, 
        ~0.210\u20130.300 during steps 75\u201395, and around 0.44\u20130.53 later 
        when stationary. 
        Values never approach 1.0, indicating contact but not firm, 
        stable placement of both hands. 
        Partial success.",
  },
  "4": {
    "number": 4,
    "name": "Align the body and box to face the destination",
    "behavior": "After getting behind the box, the agent aligns movement toward the goal; 
        the box moves straight toward the destination while the body\u2019s 
        COM velocity aligns.",
    "score": 1.0,
    "analysis": "Box velocity alignment is very high during the push 
        (reward_box_velocity_align = 0.957 at step 35, 0.999 at step 45, 0.991 at 55). 
        The body also aligns well (reward_body_velocity_align = 0.705 at step 45, 
        0.983 at step 70). 
        Behindness remains high (\u22650.95 from step 50 onward), 
        supporting correct alignment.",
  },
  "5": {
    "number": 5,
    "name": "Push the box forward while maintaining balance and steady momentum",
    "behavior": "The robot, while crouched and leaning, 
        pushes the box forward in a continuous motion toward the target. 
        The motion is steady, though posture remains low and not fully upright.",
    "score": 0.5,
    "analysis":  "Forward push quality is sustained while behind and in contact: 
        reward_forward_push = 0.279 at step 45, 0.252 at step 60, 0.269 at step 85. 
        Box motion is strongly aligned (reward_box_velocity_align \u2248 0.99 at steps 40). 
        Goal progress increases (reward_goal_distance from 0.345 at step 0). 
        Although upright is low (e.g., reward_upright = 0.091 at step 50), 
        the agent maintains functional balance sufficient to push steadily.",
  },
  "6": {
    "number": 6,
    "name": "Continue pushing, modulating force to move smoothly toward the destination",
    "behavior": "As it nears the goal, the agent reduces box speed 
        and transitions toward stop, 
        indicating some force modulation 
        rather than continuous high-speed pushing.",
    "score": 0.5,
    "analysis": "Forward push and velocity alignment taper as the box 
        approaches the destination while the stop-near-goal signal rises: 
        reward_forward_push drops from 0.077 at step 105 to 0.013 at step 120 
        and 0.004 at step 136; 
        reward_stop_near_goal increases from 0.053 at step 105 to 0.320 at step 120, 
        0.399 at step 136. 
        Meanwhile, reward_goal_distance steadily climbs to \u22480.919 by step 312, 
        showing smooth progress with deceleration near the target.",
  },
  "7": {
    "number": 7,
    "name": "Fine-position and stop within 0.1 m of the destination",
    "behavior": "The box ends up very close to the target and largely stationary,
        but the episode never registers success. 
        The agent stays pressed against the box in a crouch without achieving the final placement.",
    "score": 0.5,
    "analysis": "Despite high proximity and stopping 
        (reward_goal_distance \u2248 0.919 from step 312 onward; 
        reward_stop_near_goal \u2248 0.51\u20130.57 from steps 312\u2013499),
        the terminal success signal never triggers (reward_terminal = 0.0 at all steps). 
        This indicates the box was not within the success threshold (0.1 m) 
        and/or lacked the precise final placement required.",
  }
}
\end{verbatim}
\end{tcolorbox}

\begin{tcolorbox}[breakable, colback=pastel_light_blue!10, colframe=pastel_light_blue!70!black, title=Subtask Reflection]
\small
\begin{verbatim}
"decision": "Subtask 3 ('Place both hands on the rear face of the box") was refined.
    Because the agent consistently pushed using its torso/forearms 
    without firm bilateral hand contact. 
    Reward signals for forward motion overshadowed the contact objective,
    leading to unnatural, unintended pushing mechanics. 
    The refinement makes hand-based, bilateral contact the interface for pushing, 
    enabling reward design to gate progress and discourage non-hand pushing."

"subtasks": [
    "1. Stabilize in an upright stance.",
    "2. Walk to the box and position behind it.",
    "3. Plant both palms flat on the rear face 
        and maintain continuous bilateral hand contact as the interface for pushing.",
    "4. Align the body and box to face the destination.",
    "5. Push the box forward while maintaining balance and steady momentum.",
    "6. Continue pushing, modulating force to move the box toward the destination.",
    "7. Fine-position and stop pushing when the box is within 0.1 m 
        of the destination location."
]
\end{verbatim}
\end{tcolorbox}

\begin{tcolorbox}[breakable, colback=pastel_light_blue!10, colframe=pastel_light_blue!70!black, title=Reward Reflection]
\small
\begin{verbatim}
Revision notes
- Key issue: 
    The agent frequently pushes the box without establishing 
    firm bilateral hand contact on the rear face,
    because forward motion and progress rewards overshadow the hand contact term.
- Modifications:
    1) Gate the forward-box-motion reward by the bilateral hand contact score 
       so that forward pushing is only strongly rewarded when hands are properly placed.
    2) Reweight components to prioritize hand contact over forward speed by 
       increasing the hand contact weight and reducing the forward speed weight 
       (tightly related change to the gating).

```python
def get_reward(self) -> Tuple[float, Dict]:    
    
    ...
    
    # 3) Place both hands on the rear face of the box (close to rear point and behind plane)
    rear_offset = PACKAGE_HALF_SIZE[0]  # 0.4 m along push direction
    rear_point = package_location - dest_dir_3 * rear_offset

    def hand_contact_score(hand_pos):
        # Distance to rear point
        d = np.linalg.norm(hand_pos - rear_point)
        dist_score = np.exp(-5.0 * d)  # tight proximity rewarded

        # Behind plane alignment: dot < 0 means behind the box relative to destination
        plane_dot = float(np.dot(hand_pos - package_location, dest_dir_3))
        align_score = 1.0 / (1.0 + np.exp(10.0 * plane_dot))  # negative: ~1, positive: ~0

        return float(np.clip(dist_score * align_score, 0.0, 1.0))

    left_contact = hand_contact_score(left_hand_pos)
    right_contact = hand_contact_score(right_hand_pos)
    reward_hand_contact = float(0.5 * (left_contact + right_contact))

    # Modifictation: Bilateral contact gate, require both hands
    bilateral_contact = float(min(left_contact, right_contact))  # in [0,1]

    ...

    # 5) Push the box forward while maintaining balance and steady momentum
    box_speed_xy_vec = box_vel[:2]
    v_forward = float(np.dot(box_speed_xy_vec, dest_dir_xy))  # signed forward speed
    reward_box_forward_raw = float(1.0 - np.exp(-max(v_forward, 0.0)))
    # Modification: gate forward motion by bilateral hand contact
    reward_box_forward = float(
        np.clip(reward_box_forward_raw * bilateral_contact, 0.0, 1.0
    ))

    # 6) Progress toward destination (closeness to goal)
    dist_to_dest = float(np.linalg.norm(package_location - package_destination))
    reward_progress_raw = float(np.exp(-dist_to_dest))  
    # Modification (tightly related): gate progress by bilateral hand contact
    reward_progress = float(np.clip(reward_progress_raw * bilateral_contact, 0, 1))
    
    ...
    
    return reward
\end{verbatim}
\end{tcolorbox}

\clearpage

\section{RDA Hyperparameters}
\label{appendix:hyper_param}

This section details the hyperparameters used in our experiments (Section~\ref{sec:experiments}). We denote ManiSkill as MS and HumanoidBench as HB throughout this section.

\begin{table}[h!]
\centering
\caption{\textbf{Hyperparameters for RDA.}}
\scalebox{0.99}{
\begin{tabular}{l|l|l}
\toprule
\textbf{Category} & \textbf{Hyperparameter} & \textbf{Value} \\ \hline
\multirow{3}{*}{\textbf{Evolutionary Search}} 
    & Number of Iterations & 5 \\
    & Number of Samples per Iteration & 4 (MS), 8 (HB) \\
    & Number of Trajectories per Sample & 3 \\  \hline
\multirow{7}{*}{\textbf{Reinforcement Learning}}
    & Algorithm & Soft Actor-Critic \\ 
    & Architecture & SimbaV2 \\ 
    & Number of Environment Steps & 50M (MS), 10M (HB) \\ 
    & Number of Update Steps & 250K (MS), 625K (HB) \\ 
    & Maximum Environment Steps & Task-dependent (MS), 500 (HB) \\ 
    & Number of Training Environments & 2000 (MS), 16 (HB) \\ 
    & Discount Factor ($\gamma$) & 0.95 (MS), 0.98 (HB) \\ \hline
\multirow{3}{*}{\textbf{Agent VLM}} 
    & Type & GPT-5 (2025-08-07) \\
    & Number of Images per Query & 20 \\
    & Reasoning & Medium \\ \hline
\multirow{2}{*}{\textbf{Evaluation VLM}} 
    & Type & GPT-4.1 (2025-04-14) \\
    & Number of Images per Query & 50 \\ \hline
\end{tabular}}
\end{table}

ManiSkill supports GPU-parallelized simulation, allowing 50M environment steps per experiment. In contrast, HumanoidBench uses MuJoCo, which does not support GPU acceleration, resulting in significantly slower simulation. Accordingly, we limit training to 10M steps for HB to maintain comparable compute budgets.

All experiments were conducted on NVIDIA A100 GPUs. Each iteration in ManiSkill takes approximately 1–2 GPU hours, whereas HumanoidBench requires 6–8 GPU hours per iteration. Consequently, a complete RDA search for a single ManiSkill task requires roughly 20–40 GPU hours, while a HumanoidBench task requires 240–360 GPU hours.

\clearpage

\section{Evaluation}
\label{appendix:evaluation}

We evaluate RDA using two complementary metrics: alignment rate and success rate.

\subsection{Alignment Rate}

The alignment rate quantifies how faithfully a policy's behavior matches the intended behavior specified in the task instruction. Since this requires semantic understanding of both language and visual behavior, we employ a vision-language model as an automated evaluator.

For each trained policy, we collect 5 trajectory videos. Each video is paired with the task instruction and sent to GPT-4.1~\citep{openai2025gpt4.1}. To reduce evaluation variance, we query the VLM four times per video, yielding 20 evaluations per policy. The VLM outputs a score on a 5-point Likert scale, where higher scores indicate better alignment with the instruction. Since each task is trained with 3 random seeds, this results in 60 total evaluations per task. We report the mean score normalized to the range $[0, 1]$ by dividing by 5. We use GPT-4.1 rather than GPT-5  to mitigate potential bias between the generator and evaluator.

The system and user prompts remain fixed across all tasks and are provided below:

\begin{tcolorbox}[breakable, colback=pastel_light_blue!10, colframe=pastel_light_blue!70!black, title=VLM Evaluation System Prompt]
\small
\begin{verbatim}
You are an evaluation expert assessing whether a video accurately 
follows a given task instruction.
Your goal is to judge how well the video execution matches the intended task 
on a 5-point scale.

## Instruction
1. Understand the task
Use the following inputs to determine what behavior is expected:
- Task description: defines the goal and intended sequence of actions.
- Video: shows the actual performance of the task.

2. Evaluate performance accuracy
- Compare what happens in the video with what the task description specifies.
- Focus on goal completion, action correctness, and order of steps.

3. Scoring criteria (1–5 scale)
- (1) Completely incorrect: Unrelated or failed attempt; goal not achieved.
- (2) Mostly incorrect: Limited or wrong attempt; major steps missing.
- (3) Partially correct: Some correct actions, but key errors or omissions.
- (4) Mostly correct: Follows most steps; minor mistakes; goal achieved.
- (5) Fully correct: All actions and goals performed as intended.

## Output
Return only a numerical integer score (1–5).
\end{verbatim}
\end{tcolorbox}

\begin{tcolorbox}[breakable, colback=pastel_light_blue!10, colframe=pastel_light_blue!70!black, title=VLM Evaluation User Prompt]
\small
\begin{verbatim}
You are given the input to evaluate how accurately a video follows a task instruction:
- Task instruction: outlines the goal and intended agent behavior.
- Video: shows the actual behavior of the agent.

Using this information, assign a numerical score (1–5) based on 
how accurately the actions in the video match the task instruction.

## Task Instruction
{task_instruction}

## Video
{video}
\end{verbatim}
\end{tcolorbox}

\subsection{Success Rate}

The success rate measures task completion using binary success indicators provided by each benchmark. Unlike alignment rate, this metric evaluates only whether the task goal is achieved, not whether the policy follows the intended behavior specified in the instruction.

For ManiSkill, we adopt the official success-checking functions; for HumanoidBench, we use the terminal reward bonus as a binary indicator of task completion.

\section{Baseline Details}

We provide implementation details for all baseline methods. Unless otherwise specified, all baselines use identical training configurations as RDA to ensure fair comparison: the same RL algorithm (SAC~\citep{haarnoja2018soft} with SimbaV2~\citep{lee2025hyperspherical}), number of environment steps, and candidates.

\subsection{Human Design}

We evaluate two variants of human-designed reward functions to establish lower and upper bounds on what can be achieved without LLM-based reward generation.

\textbf{Sparse Reward.} The sparse baseline uses the binary success indicator provided by each benchmark as the reward signal: the agent receives a reward of 1 when the task is successfully completed and 0 otherwise. This serves as a lower bound on performance and uses the same computational budget as RDA.

\textbf{Dense Reward.} The dense baseline uses hand-crafted reward functions provided by the benchmark developers. For ManiSkill, we use the official dense reward functions provided in the benchmark code, which typically combine distance-based shaping with success bonuses. For HumanoidBench, we use the dense reward functions defined in the original benchmark paper~\citep{sferrazza2024humanoidbench}, which decompose the task into multiple reward components such as object tracking, balance penalties, and goal proximity. These dense rewards represent expert human engineering effort and serve as a strong baseline for reward design quality.

\subsection{Eureka}

We implement Eureka~\citep{ma2023eureka} as our primary baseline for LLM-based reward generation. To ensure fair comparison, we match the configuration as closely as possible to RDA: GPT-5 with medium reasoning effort as the VLM backbone with identical training budgets and task instructions. The key difference between Eureka and RDA is the reflection mechanism: Eureka reflects only on numerical reward statistics (mean, minimum, and maximum of each reward component), whereas RDA performs visual trajectory analysis to diagnose failure modes and assess subtask completion.

The system and user prompts used for Eureka are provided below:
\begin{tcolorbox}[breakable, colback=pastel_light_blue!10, colframe=pastel_light_blue!70!black, title=Reward Generation System Prompt]
\small
\begin{verbatim}
You are a reward engineering expert assisting reinforcement learning agents.
Your job is to design a reward function that maximizes the likelihood of task success.

## Instructions
1. Understand the task
Use the following inputs to understand the task:
- task instruction: outlines the goal and intended agent behavior.
- Environment code: includes the agent's observation and **success condition**.

2. Implement the Reward Function
- Implement the function inside the environment class using the {reward_signature}.
- Use only existing attributes and methods from the provided environment code.
- Do not create or assume new fields beyond those defined.

3. Reward Design
- Shape rewards to encourage continuous progress toward success.
- Provide a large, one-time terminal reward when success is achieved.

4. Normalization and scaling
- Normalize each reward component into a bounded range within [0,1] before combining.
- Shaping functions (e.g. exp, tanh) may be used as part of producing this bounded form.
- After normalization, multiply each component by its explicit weight.
- The final reward is the sum of (weight × normalized_component) across all components.

5. Logging
- Define a dictionary named self.reward_info inside the function.
- Store each component with keys prefixed by "reward_" (e.g., "reward_distance")

## Output
Return only a single Python code block implementing the function:
```python
# your code here
```
\end{verbatim}
\end{tcolorbox}

\begin{tcolorbox}[breakable, colback=pastel_light_blue!10, colframe=pastel_light_blue!70!black, title=Reward Generation User Prompt]
\small
\begin{verbatim}
You are given the following input to design a reward function for an RL agent:
- task instruction: outlines the goal and intended agent behavior.
- Environment code: includes agent's observation and **success condition**.

Using this information, implement a reward function leading the agent to succeed the task.

## Task Instruction
{task_instruction}

## Environment Code
{env_code}
\end{verbatim}
\end{tcolorbox}

\begin{tcolorbox}[breakable, colback=pastel_light_blue!10, colframe=pastel_light_blue!70!black, title=Reward Reflection System Prompt]
\small
\begin{verbatim}
You are a reward engineering expert assisting reinforcement learning agents.
Your goal is to revise an existing reward function to maximize task success 
using analysis and results.

## Instructions

1. Understand the context
Use the following inputs to understand the context of the task:
- Task description: outlines the goal and intended agent behavior.
- Environment code: includes agent's observation and **success condition**.
- Previous reward function and its analysis:
  - reward function code.
  - task success rate (from the model’s reflection).
  - reward statistics in the form of dict(step_0: value, step_1: value, ..., step_N: value)

2. Revise the Reward Function
- Revise the reward function inside the environment class using the exact signature:
```
{reward_signature}.
```
- Use only existing attributes and methods from the provided environment code.
- Do not create or assume new fields beyond those defined.

3. Revision Strategy
- Make key changes that directly address the reported failure(s).
- You may re-weight existing components as part of the revision.

4. Reward Design
- Shape rewards to encourage continuous progress toward success.
- Provide a large, one-time terminal reward when success is achieved..

5. Normalization and scaling
- Normalize each reward component into a bounded range within [0,1] before combining.
- Shaping functions (e.g. exp, tanh) may be used as part of producing this bounded form.
- After normalization, multiply each component by its explicit weight.
- The final reward is the sum of (weight × normalized_component) across all components.

6. Logging
- Define a dictionary named self.reward_info inside the function.
- Store each component with keys prefixed by "reward_" (e.g., "reward_distance").

## Output

1. Revision notes
Briefly summarize what was changed and which failures these changes target.

2. Revised Reward Function
Return a single Python code block implementing the function:
```python
# your code here
```
\end{verbatim}
\end{tcolorbox}

\begin{tcolorbox}[breakable, colback=pastel_light_blue!10, colframe=pastel_light_blue!70!black, title=Reward Reflection User Prompt]
\small
\begin{verbatim}
You are given the following information to revise a reward function for an RL agent:
- Task description: outlines the goal and intended agent behavior.
- Environment code: includes agent's observation and **success condition**.
- Previous reward function and its analysis, including:
  - reward function code.
  - task success rate (from the model’s reflection).
  - reward statistics in the form of dict(step_0: value, ..., step_N: value)

Revise the reward function by introducing one modification that addresses key failures.
The revised reward should improve the likelihood of task success.

## Task Instruction
{task_instruction}

## Environment Code
{env_code}

## Previous Reward Function and its Analysis
### Previous Reward Function
{reward_function}

### Success Rate
{self_task_success}

### Reward Statistics
{reward_statistics}
\end{verbatim}
\end{tcolorbox}

\clearpage

\section{Full Results}
\label{appendix:full_result}

We report per-environment results for RDA and baseline methods.

\subsection{Alignment Rate}

\begin{figure}[h]
\begin{center}
\includegraphics[width=.98\textwidth]{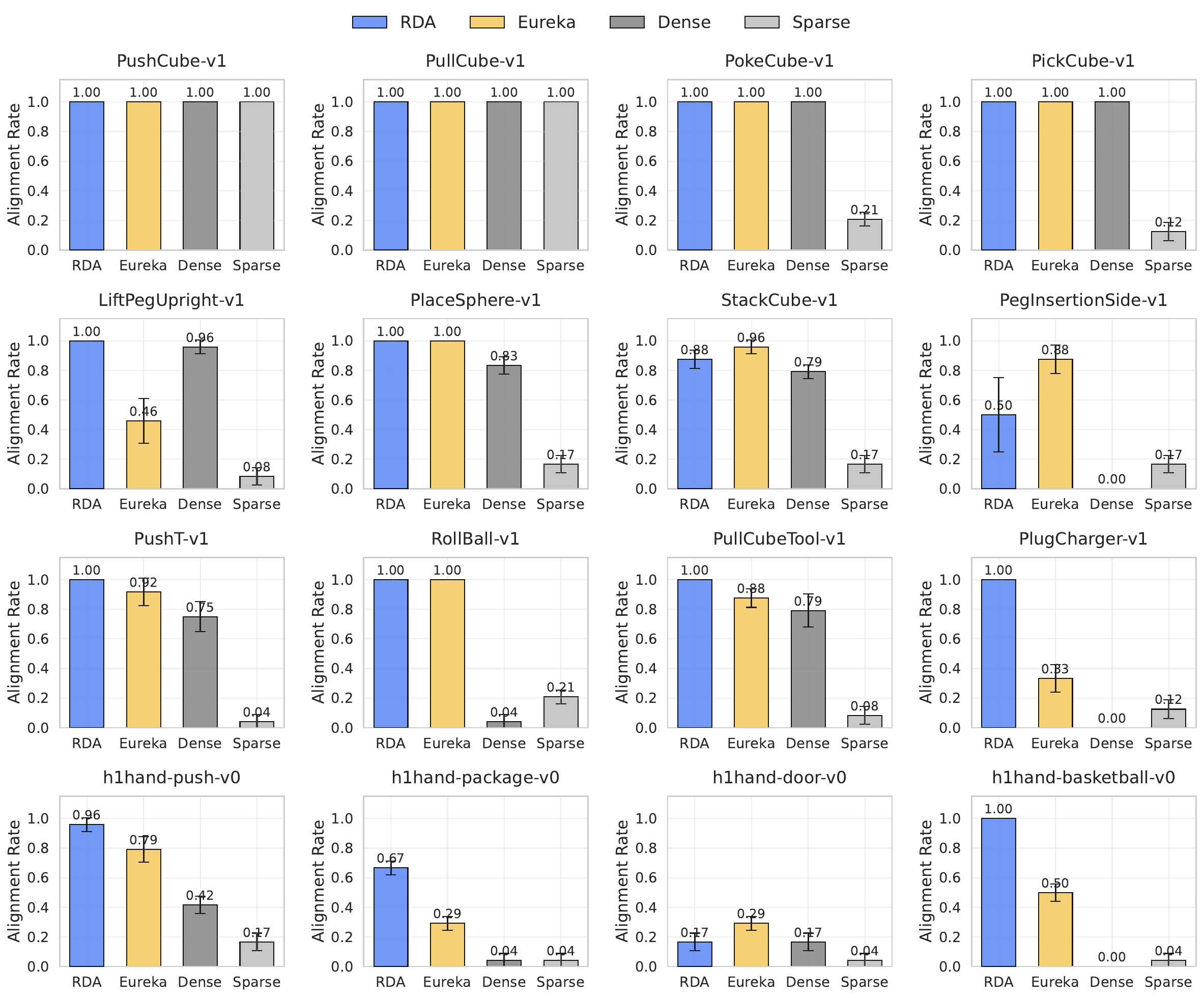}
\end{center}
\caption{\textbf{Alignment Rater.} Per-environment task alignment rate based on VLM-based evaluation. Higher values indicate better alignment between the generated trajectories and task.}
\vspace{-2mm}
\label{figure:quantitative_alignment}
\end{figure}

\clearpage

\subsection{Success Rate}

\begin{figure}[h]
\begin{center}
\includegraphics[width=.98\textwidth]{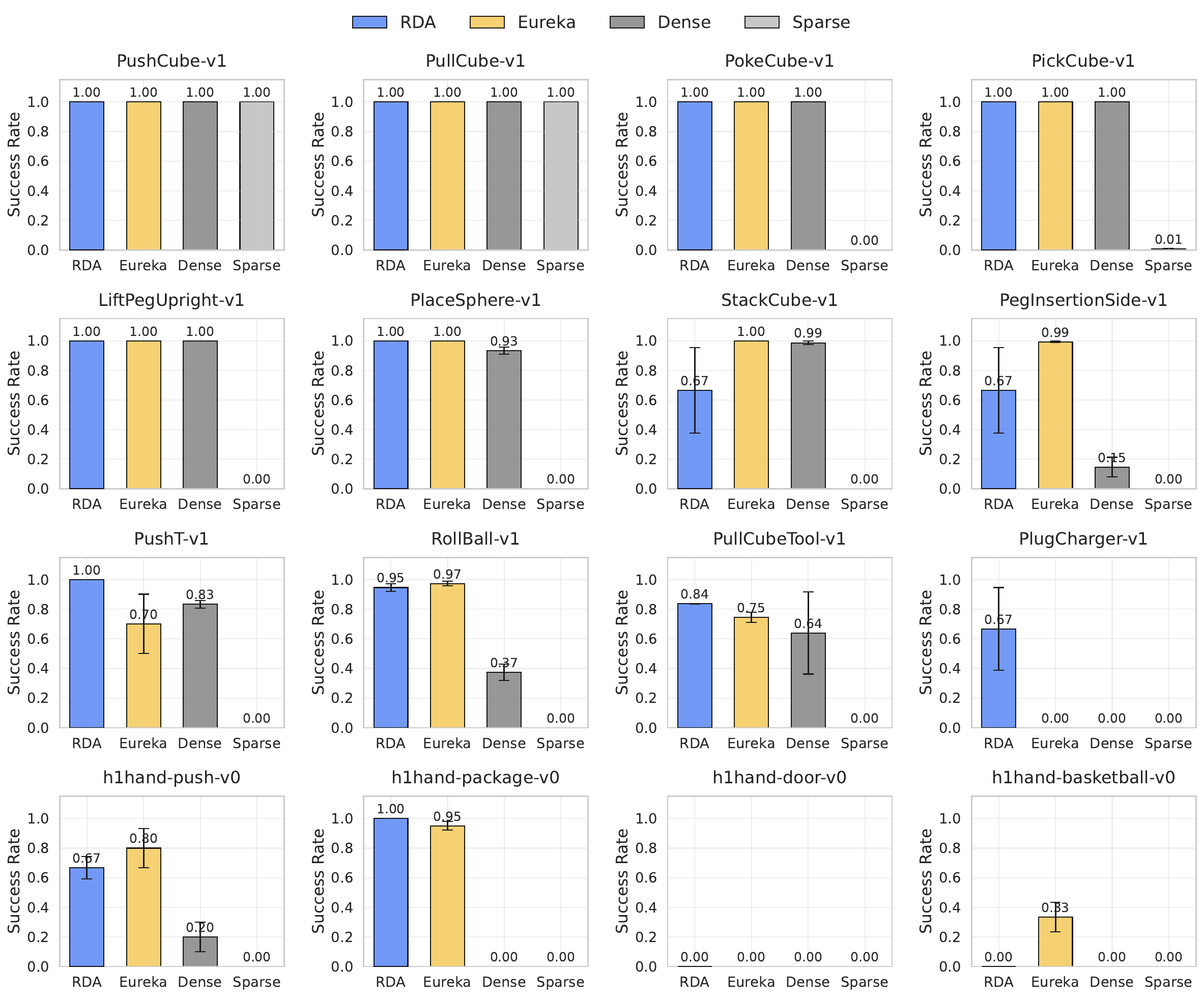}
\end{center}
\caption{\textbf{Success Rate.} Per-environment binary success rates derived from human-coded evaluation. This metric reflects whether the agent successfully completes the task, regardless of behavioral quality.}
\vspace{-2mm}
\label{figure:quantitative_success}
\end{figure}

\end{document}